\documentclass{article}
\PassOptionsToPackage{numbers,square,sort&compress}{natbib} 
\usepackage[preprint]{nips}

\usepackage[utf8]{inputenc}
\usepackage[T1]{fontenc}
\usepackage{hyperref}
\usepackage{url}
\usepackage{booktabs}
\usepackage{amsfonts}
\usepackage{nicefrac}
\usepackage{microtype}
\usepackage{xcolor}
\definecolor{dynblue}{RGB}{50,100,180}
\definecolor{dynorange}{RGB}{210,110,40}
\hypersetup{
  colorlinks=true,
  linkcolor=dynblue,
  citecolor=dynblue,
  urlcolor=dynblue
}
\usepackage{graphicx}
\usepackage{amsmath}
\usepackage{multirow}
\usepackage{enumitem}
\usepackage{makecell}

\title{\textcolor{dynblue}{Dynamic}\textcolor{dynorange}{Manip}: Enabling Dynamic Manipulation from a Single Static Demonstration}

\author{%
  {\normalfont Haoran Liao$^{1,*}$ \quad Pengyue Wang$^{1,*}$ \quad Shuoyu Chen$^{1,*}$ \quad Kehan Cheng$^{1}$} \\
  {\normalfont Xuhang Chen$^{1}$ \quad Yuhao Lin$^{1}$ \quad Mu Lin$^{1}$ \quad Zhizhao Liang$^{1}$} \\
  {\normalfont Xiaoyi Fan$^{4}$ \quad Chengyi Xing$^{2}$ \quad Dan Niu$^{3}$ \quad Yi-Lin Wei$^{1}$ \quad Wei-Shi Zheng$^{1}$} \\
  \small{$^{1}$Sun Yat-sen University} \quad  $^{2}$Stanford University \quad $^{3}$Southeast University \quad $^{4}$Jiangxing Intelligence\\
    \small{
      Project page:
      \href{https://liaohr9.github.io/DynamicManip/}
           {\textcolor{dynblue}{https://liaohr9.github.io/DynamicManip/}}
      \quad $^{*}$Equal contribution.
    }
}

\begin{document}

\maketitle

\begin{figure*}[ht]
  \centering
  \vspace{-1cm}
  \includegraphics[width=1\textwidth]{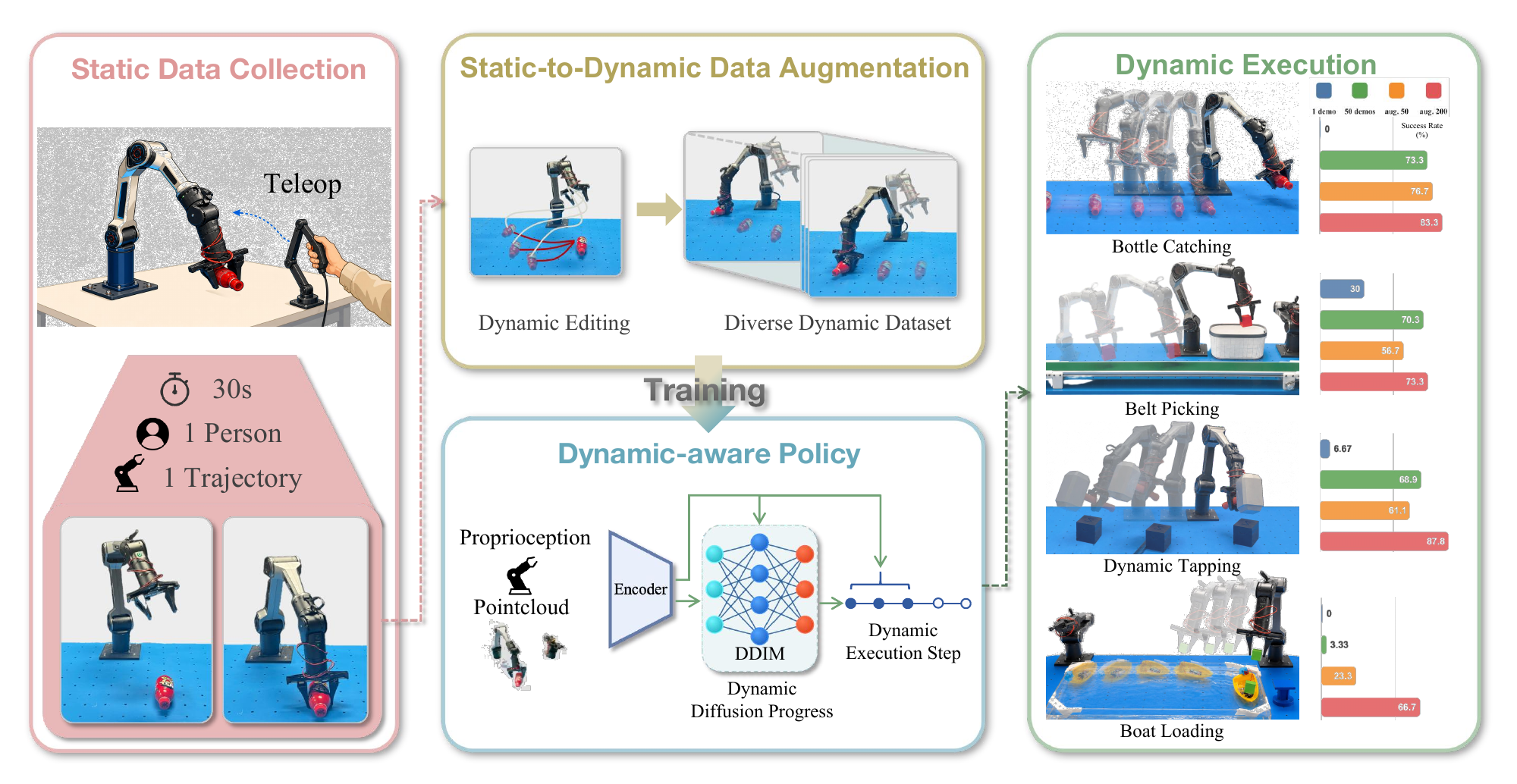}

\caption{\textbf{DynamicManip} generates diverse dynamic manipulation data from a single static task demonstration in the real world. Leveraging the augmented data, our dynamic-aware adaptive policy enables generalizable dynamic manipulation with efficient inference.}
  \label{fig:teaser}
  \vspace{-0.5em}
\end{figure*}


\begin{abstract}
Dynamic manipulation is a critical capability for robots operating in complex and dynamic environments, where robots must interact with objects that are moving or require rapid adjustments. 
However, learning models for dynamic manipulation tasks face two major challenges: (1) the combinatorial complexity of dynamic scenarios leads to substantial data requirements, and (2) rapid variations in dynamics require real-time and accurate policy execution.
In this paper, we propose \textbf{DynamicManip} to address these challenges through an efficient data augmentation pipeline and a low-latency imitation policy. 
We first propose a static-to-dynamic augmentation pipeline that synthesizes diverse dynamic manipulation demonstrations from a single static demonstration. 
Second, we introduce a dynamic-aware adaptive policy that adjusts its inference frequency according to task dynamics, enabling responsive and effective dynamic manipulation.
Third, we build a dynamic manipulation benchmark, which includes diverse dynamic tasks with an automatic evaluation system for scalable and consistent assessment.
Extensive experiments in both simulation and the real world demonstrate that DynamicManip not only provides significant improvements in data efficiency but also achieves better performance in dynamic manipulation tasks, with a mean success rate 18.4 percentage points higher and policy-query latency 32.9\% lower.
\end{abstract}

\section{Introduction}
\label{sec:introduction}

A long-term goal in robotics is to deploy robots in everyday human environments~\cite{argall2009survey,billard2019trends,wei2025cyclemanip}, where they can assist people and integrate into our daily workflows.
To realize this vision, robots must handle dynamic tasks in which the target objects are in motion, such as catching a flying ball or stopping a glass about to slip off the table~\cite{zhang2025dynamic}.
These tasks require the ability of a robot to continuously perceive, track, and respond to objects in motion through closed-loop perception-action cycles. 

However, dynamic manipulation remains significantly challenging compared to static tasks.
Two fundamental challenges stand in the way~\cite{xie2026dynamicvla}.
First, collecting data for dynamic tasks is inherently difficult, owing to the combinatorial explosion of object velocities, trajectories, and robot response timings ~\cite{lin2025typetele,zhao2026hail}.
Second, dynamic manipulation demands real-time model inference to cope with rapid state variations~\cite{rdt2, zhang2025catch,liao2025delay,black2025real,sendai2025leave}. Traditional methods typically train on quasi-static datasets and trade inference speed for precision, making them less suitable for dynamic tasks \cite{chi2025dp}.

To overcome these limitations, we present \textbf{DynamicManip}, which addresses the challenges of dynamic manipulation learning through 
(1) a static-to-dynamic data augmentation pipeline that synthesizes diverse dynamic demonstrations from a single static trajectory while preserving the robot behavior patterns and underlying physical consistency; 
and (2) a dynamic-aware adaptive policy that 
adapts its inference rate to evolving manipulation dynamics, improving responsiveness and efficiency while maintaining or improving task success in our experiments.

For data augmentation, we introduce a static-to-dynamic pipeline that flexibly modifies object dynamics and robot motion. Specifically, the pipeline 
first reconstructs the manipulation environment, allowing point cloud rendering of both objects and the robot under arbitrary configurations. 
It then edits object trajectories and performs motion planning to synthesize paired observations and actions that preserve geometric feasibility and geometrically feasible robot-object interaction trajectories.

Based on the augmented data, DynamicManip improves responsiveness by adaptively adjusting its inference strategy across manipulation stages. 
Specifically, we define a dynamic-stage indicator that specifies whether each stage requires high or low computational effort. 
The dynamic-stage indicator is automatically annotated by our augmentation pipeline using heuristics derived from robot-object interaction dynamics. 
A lightweight auxiliary head is then trained to predict the indicator at test time, enabling the policy to adjust diffusion denoising steps or action-chunk execution adaptively.

To support our framework, we introduce the DynamicManip Benchmark.
The benchmark includes diverse dynamic manipulation tasks and an automatic
evaluation pipeline for systematic experiments.
Built on the RoboTwin 2.0 platform~\cite{chen2025robotwin}, our benchmark provides highly configurable dynamic task environments for data collection and evaluation, with an automatic evaluation pipeline that measures task success rates and response latency.
This enables a systematic assessment of dynamic manipulation policies.

Comprehensive experiments in both simulation and the real world demonstrate that DynamicManip effectively scales a single static human demonstration into a diverse dynamic dataset, reducing the amount of repeated dynamic teleoperation required for the evaluated tasks. Moreover, our proposed policy achieves substantially faster execution while maintaining high manipulation precision, enabling responsive closed-loop control in dynamic scenarios and improving task success rates across diverse dynamic manipulation tasks.


\section{Related Work}
\label{sec:related}

\subsection{Robot Manipulation}
\label{sec:related:manipulation}

Robot manipulation has long been a central research area in robotics and is widely viewed as a key pathway toward general-purpose embodied intelligence~\cite{bai2025towards, wei2025cyclemanip, billard2019trends}. Recent progress in Imitation Learning (IL)~\cite{zhao2023act, james2020rlbench, liu2023libero, brohan2022rt, zitkovich2023rt}
and Vision-Language-Action (VLA) models~\cite{black2024pi0, intelligence2025pi05, intelligence2026pi, liu2024rdt, kim2024openvla, bjorck2025gr00t} has shown strong capability in learning action distributions from large-scale expert demonstrations and predicting action sequences conditioned on observations, while complementary efforts have explored specialized manipulation capabilities such as language-guided dexterous grasp generation~\cite{wei2024grasp} and foundation-model-based grasp transfer with force-aware adaptation~\cite{wei2025omnidexgrasp}.
Despite this progress, most existing manipulation methods and benchmarks are primarily evaluated in static or slowly changing settings, where target objects remain stationary or change only slightly during manipulation.
In these scenarios, data collection is often more manageable than in dynamic manipulation, since demonstrations can be obtained through standard teleoperation without covering diverse object motions.
Moreover, policies can often tolerate moderate inference latency because the scene evolves slowly during action execution.

\subsection{Dynamic Tasks}
\label{sec:related:dynamic}

Dynamic manipulation is pervasive in everyday and industrial environments, from catching a slipping glass to sorting products on moving conveyors. 
Such tasks require a robot to continuously track and respond to objects whose states evolve rapidly or unpredictably, demanding rapid perception, predictive planning, and low-latency action generation within closed-loop perception-action cycles. 
They also impose a higher data burden, as demonstrations must capture diverse object motions, interaction timings, and robot-object temporal alignments.
Despite its importance, research on general dynamic manipulation under uncertain motion and fine-grained contact constraints remains relatively limited.
Existing methods largely focus on static settings or structured scenarios with predictable motion, such as DBC-TFP~\cite{zhang2025dynamic} and GEM~\cite{li2025gem} in conveyor-like environments. Concurrent VLA approaches, including RDT-2~\cite{rdt2}, RTVLA~\cite{ma2025running}, and VLASH~\cite{tang2025vlash}, demonstrate real-time interaction with fast-moving targets but operate with large contact margins and lack precise manipulation. Although works such as DynamicVLA~\cite{xie2026dynamicvla,kim2014catching,akinola2021dynamic} attempt to address this problem, a significant gap remains between current capabilities and the demands of real-world dynamic manipulation.

\subsection{Data Collection and Augmentation}
\label{sec:related:augmentation}

High-quality and diverse demonstration data are critical for learning robust manipulation policies\cite{belkhale2023data,khazatsky2024droid,lin2025data,zhu2025unified,lyu2026lda,xiao2025robot}, yet collecting such data remains expensive\cite{fang2023rh20t,o2024open,geng2025roboverse,lin2025typetele}, especially for dynamic tasks with large combinatorial variation\cite{gao2026dreamdojo,ye2026world,lyu2025dywa}. 
Data augmentation has therefore emerged as a promising direction to scale limited demonstrations\cite{yu2025real2render2real}.
Existing approaches can be broadly categorized into several groups. 2D visual augmentation methods (e.g., GenAug~\cite{chen2023genaug,mandlekar2023mimicgen}) modify appearance through image-level transformations but do not enforce 3D consistency. Implicit 3D reconstruction methods (e.g., RoboSplat~\cite{yang2025novel}) improve visual fidelity via neural scene representations, yet typically do not adapt action trajectories accordingly. Geometric augmentation approaches (e.g., RoCoDA~\cite{ameperosa2025rocoda}) apply SE(3)-equivariant transformations in state space but are largely limited to rigid global transformations. Simulation-based methods (e.g., FAR-Dex~\cite{bai2026far}) can generate physically plausible trajectories but may suffer from sim-to-real discrepancies. Grasp-level synthesis methods such as BiDexGrasp~\cite{lin2026bidexgrasp} generate physically feasible bimanual grasp configurations across diverse object geometries and sizes, but do not model temporally coordinated robot trajectories under evolving object motion. At the trajectory level, more recent work such as DemoGen~\cite{xue2025demogen} explores trajectory editing for tabletop tasks, but only generates quasi-static variations and does not address the temporal dynamics critical for dynamic manipulation.
In contrast, our method turns minimal static demonstrations into geometrically feasible and physically consistent dynamic demonstration data by jointly modifying object motion and adapting the corresponding robot actions.

\section{Method}
\label{sec:method}

\subsection{Overview}
\label{sec:method:overview}

We study dynamic manipulation from extremely limited demonstrations, where the robot must learn to interact with moving objects while maintaining low perception-action latency. 
DynamicManip addresses this problem by coupling data generation with adaptive policy inference:
a static-to-dynamic augmentation pipeline converts a single static demonstration into diverse dynamic episodes with timestep-level interaction-stage labels,
and a dynamic-aware imitation learning policy uses these labels to learn stage prediction and automatically adjust its inference behavior from online observations during deployment.

\begin{figure}
  \centering
  \includegraphics[trim=0mm 10mm 0mm 0mm, clip, width=0.9\textwidth]{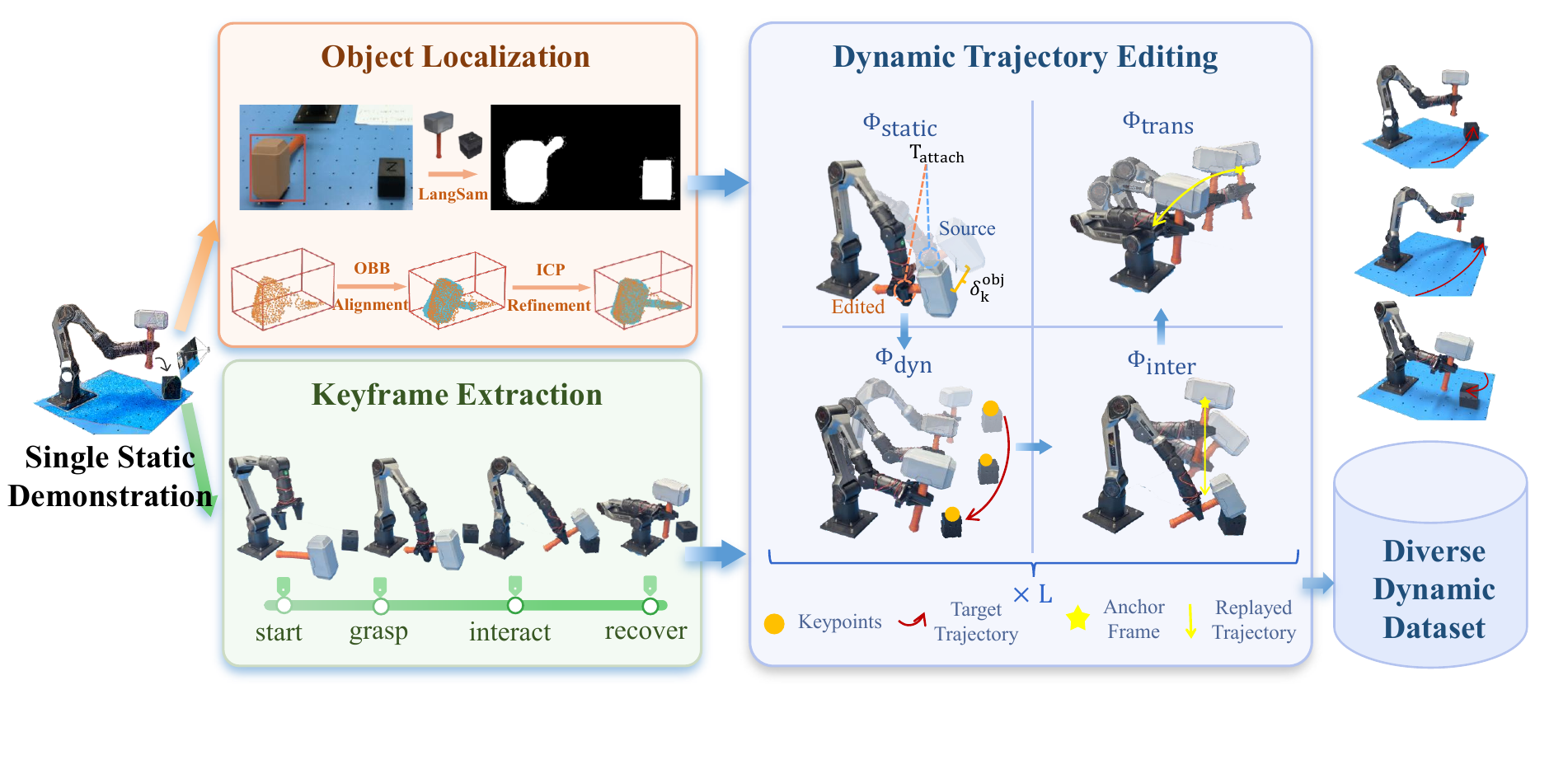}
    \vspace{-1.3em}
    \caption{The pipeline of our static-to-dynamic data augmentation. From a single static demonstration, we localize the manipulation object and target and reconstruct their geometry. We then extract action-based keyframes, where labels such as start, grasp, interact, and recover only serve as intuitive annotations for segmenting the source motion into reusable pieces. Dynamic trajectory editing then organizes stages such as Dynamic Target Alignment ($\Phi_{\mathrm{dyn}}$) to synthesize diverse trajectories, which are generated through motion planning and replaying local interaction segments.}
  \label{fig:augmentation_pipeline}
    \vspace{-1em}
\end{figure}

\subsection{Static-to-Dynamic Data Augmentation}
\label{sec:method:augmentation}

Our augmentation pipeline starts from the observation that a static demonstration already specifies \emph{how} the robot should interact with the object, while dynamic variations arise mainly from \emph{where} and \emph{when} the interaction occurs.
Therefore, instead of collecting dynamic demonstrations directly, we preserve the local interaction patterns in the source episode and edit the object motions, phase anchors, and transition targets to synthesize new dynamic episodes.
Given a single source episode $\mathcal{E}_{\text{src}} = \{(\mathbf{p}_t, \mathbf{e}_t, \mathbf{q}_t)\}_{t=0}^{T}$, where $\mathbf{p}_t \in \mathbb{R}^{N \times 3}$ is the scene point cloud, $\mathbf{e}_t \in SE(3)$ is the end-effector pose, and $\mathbf{q}_t \in \mathbb{R}^{d}$ is the robot state, our pipeline generates $K$ augmented dynamic episodes through geometric reconstruction, keyframe extraction, and dynamic trajectory editing.

\paragraph{Geometric reconstruction.}
The first step is to build an editable geometric representation for subsequent trajectory editing.
Directly transforming the recorded point clouds is unreliable, because single-view observations are incomplete and view-dependent: spatial edits can expose object surfaces that were occluded in the source demonstration.
Therefore, we reconstruct both the manipulated object and the robot geometry before trajectory editing.
For the object, we generate a CAD mesh from RGB images and align it to the observed object point cloud $\mathcal{P}_o$ using OBB-based coarse initialization followed by multi-resolution ICP refinement~\cite{icp}. The aligned mesh is then resampled into a canonical object point cloud, which can be transformed coherently under arbitrary spatial edits. 
For the robot, the geometry is obtained from the known URDF model rather than reconstructed from partial observations. Given the URDF model and joint configuration $\mathbf{q}_t$, we compute the pose of each robot link via forward kinematics and sample the transformed link meshes to obtain the robot point cloud $\mathcal{P}^{\mathrm{robot}}_t$.
Together, the reconstructed object point cloud and the rendered robot point cloud provide a complete and consistent scene representation for subsequent steps.

\paragraph{Keyframe Extraction.}
The reconstructed geometry makes the scene editable, but dynamic trajectory synthesis also requires knowing which parts of the source motion can be reused and where they should be anchored.
Therefore, we decompose the static demonstration into reusable phase segments using sparse keyframe annotations.
These keyframes mark the boundaries between approach, interaction, and disengagement phases, and provide anchor frames for transforming and recomposing the corresponding motion segments during subsequent trajectory editing.
This sparse annotation is lightweight, but it preserves the temporal structure needed to reuse the source demonstration across diverse dynamic episodes.

\paragraph{Dynamic trajectory editing.}
Given the editable scene representation and the annotated phase segments, we synthesize dynamic episodes by reusing the source interaction segments under newly sampled object poses, object motions, and transition targets.
For the $k$-th augmented episode, we sample a set of task-specific augmentation parameters $\xi_k \sim \mathcal{D}$, which specify object poses, object motion trajectories, phase anchors, and transition goals.
The sampled parameters determine how each reusable segment is transformed, planned, or replayed in the new dynamic scene.
Guided by $\xi_k$, we instantiate and compose four phase operators to generate a complete dynamic trajectory:

\noindent\textit{Static Object Acquisition} $(\mathbf{\Phi}_{\mathrm{static}})$
Many dynamic tasks begin with a static preparation stage, where the robot grasps or picks up an object before interacting with a moving target.
Since the object is static, this segment can be augmented by applying a rigid transformation $T_k^{\mathrm{static}}\in SE(3)$ to the entire phase:
$\mathbf{p}'_t=T_k^{\mathrm{static}}\mathbf{p}_t,\ \mathbf{e}'_t=T_k^{\mathrm{static}}\mathbf{e}_t$.
This transformation increases spatial diversity while preserving the relative motion pattern between the robot and object.

\noindent\textit{Dynamic Target Alignment} $(\mathbf{\Phi}_{\mathrm{dyn}})$.
In this phase, the robot aligns its end-effector with a moving object or a predicted target before executing the interaction.
Given a synthesized object motion $T_{\mathrm{obj}}^{k}(t)$, we first compute a time-varying interaction target $\mathbf{g}^{k}(t)$ from the task-specific prediction rule.
For example, $\mathbf{g}^{k}(t)$ may correspond to an object-frame relative pose or a predicted landing point derived from a parabolic object trajectory.
A motion planner then generates a smooth trajectory to reach these target-conditioned poses while satisfying robot constraints.
This planning-based construction allows the same static demonstration to be reused across diverse object motions while keeping the generated alignment motions smooth and kinematically feasible.

\noindent\textit{Contact Interaction Execution} $(\mathbf{\Phi}_{\mathbf{inter}})$.
In this phase, the robot executes the short contact-rich motion that is difficult to specify with motion planning alone, such as closing the gripper during dynamic grasping or striking a target with a hammer.
Let $A$ be the anchor frame of the source interaction segment, such as the object or end-effector pose at the contact keyframe. 
For the $k$-th augmented episode, the edited trajectory defines a new anchor frame $A_k$, and we transfer the local interaction motion by $\mathbf{e}'_t = A_k A^{-1}\mathbf{e}_t$.
This anchor-based replay preserves the fine local motion pattern while allowing the same interaction to occur at different object positions and orientations.

\noindent\textit{Pose-Conditioned Transition} $(\mathbf{\Phi}_{\mathbf{trans}})$.
In this phase, the robot moves between reusable motion segments, such as transporting an object, reaching the next interaction pose, or returning to the reset configuration.
Given a recorded keyframe pose $\mathbf{e}_{\mathrm{key}}$ and a pose offset $T_k^{\mathrm{trans}}\in SE(3)$, we define the transformed target pose as
$\mathbf{e}_{\mathrm{key}}^{k} = T_k^{\mathrm{trans}}\mathbf{e}_{\mathrm{key}}$.
A motion planner then generates a feasible trajectory from the current state to $\mathbf{e}_{\mathrm{key}}^{k}$. 
This transition phase connects the edited interaction segments into a continuous executable trajectory.

\noindent\textbf{Composing trajectories with stage labels.}
The four phase operators can be composed to instantiate complete dynamic manipulation trajectories.
For example, a common trajectory structure is 
$\Phi_{\mathrm{static}}
\rightarrow
(\Phi_{\mathrm{dyn}}
\rightarrow
\Phi_{\mathrm{inter}}
\rightarrow
\Phi_{\mathrm{trans}})^L$,
where $L$ denotes the number of interaction cycles.
Different tasks can be generated by reordering, skipping, or repeating these operators according to the task structure.
During composition, each timestep inherits the identity of the phase operator that generates it, which provides an automatic stage label without additional annotation.

By sampling different augmentation parameters $\xi_k$, the pipeline produces diverse dynamic trajectories from a single keyframe-annotated static demonstration.
The resulting stage-labeled trajectories are used in the next section
to train a dynamic-aware policy.

\subsection{Dynamic-aware Imitation Learning Policy}
\label{sec:method:model}

\begin{figure}
    \centering
    \includegraphics[width=\linewidth]{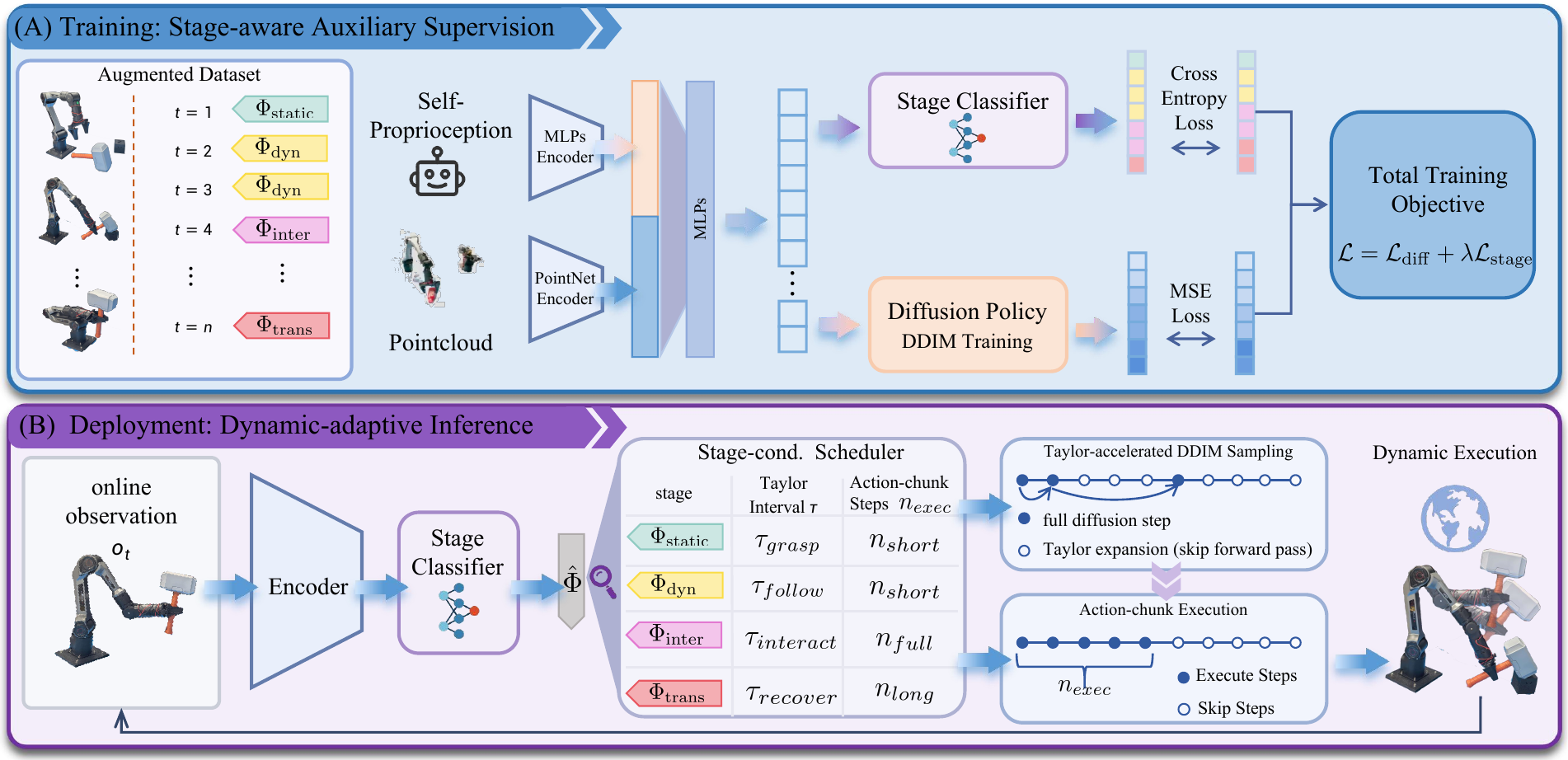}
    \caption{\textbf{Overview of the dynamic-aware adaptive policy.} During training, stage labels automatically extracted from augmented trajectories provide auxiliary supervision for stage prediction. During deployment, the predicted stage is used to adaptively regulate the policy inference frequency according to the current task dynamics.}
    \label{fig:enter-label}
    \vspace{-0.5em}
\end{figure}

Based on the augmented data, we train a dynamic-aware imitation learning policy that is both accurate and responsive for dynamic manipulation. Specifically, we introduce a stage-aware inference mechanism, which automatically applies inference strategies based on the robot-object interaction stage. 
During training, a lightweight auxiliary head learns to predict the interaction stage from visual observations.
During deployment, the predicted stage is used to select an appropriate inference strategy according to the current task dynamics.

\paragraph{Dynamic-stage Awareness Learning}
To make the policy aware of its interaction stage, we supervise it with the stage labels automatically produced by the augmentation pipeline. 
Specifically, every augmented trajectory carries a stage label $\ell_t \in \{\Phi_{\mathrm{static}}, \Phi_{\mathrm{dyn}}, \Phi_{\mathrm{inter}}, \Phi_{\mathrm{trans}}\}$ for each timestep~$t$, reflecting the interaction logic that produced it. We attach a lightweight classification head $h_\phi$ to the shared observation encoder. Given encoded features $\mathbf{f}_t$, the head predicts a stage distribution $\hat{\mathbf{p}}_t = h_\phi(\mathbf{f}_t)$ over the four stages, supervised by a cross-entropy loss:
\begin{equation}
\mathcal{L}_{\text{stage}} = -\frac{1}{T}\sum_{t=1}^{T} \log \hat{p}_t(\ell_t).
\end{equation}
The total training objective combines the diffusion loss with this auxiliary term:
\begin{equation}
\mathcal{L} = \mathcal{L}_{\text{diff}} + \lambda\,\mathcal{L}_{\text{stage}},
\end{equation}
where $\lambda$ balances action fidelity and stage awareness. This auxiliary objective encourages the encoder to encode the current stage, which helps the policy adaptively adjust its inference strategy.

\paragraph{Dynamic-adaptive inference.}
At deployment, the trained stage classifier predicts the current interaction stage at each at each policy-query cycle: $\hat{\ell}_t = \arg\max_k \hat{p}_t(k)$. 
Conditioned on this stage label, the policy dynamically adjusts its computational allocation, including the denoising schedule and the action-chunk execution length. 

The denoising schedule controls how frequently the diffusion model performs full forward computation during DDIM sampling~\cite{song2020ddim}. 
For intermediate denoising steps where full computation is skipped, we adopt a Taylor-based acceleration mechanism~\cite{liu2025reusing} that treats each residual block output $\mathbf{y}(s)$ as a smooth function of the diffusion step index~$s$ and approximates it via a Taylor expansion around a recently computed anchor step $s_0$:
\begin{equation}
\mathbf{y}(s) \approx \sum_{k=0}^{K} \frac{1}{k!}\, \mathbf{y}^{(k)}(s_0)\, (s - s_0)^k,
\end{equation}
where the derivatives $\mathbf{y}^{(k)}(s_0)$ are estimated via backward finite differences between consecutive full-computation steps (Appendix~\ref{sec:appendix:inference}). By skipping expensive forward passes at intermediate denoising steps, this approximation reduces inference latency while maintaining action accuracy.

In addition to denoising acceleration, we adapt the action-chunk execution length, which controls how many predicted actions are executed before the next policy query.
Shorter chunks provide more frequent feedback, whereas longer chunks reduce inference calls and improve motion continuity. 

The predicted stage $\hat{\ell}_t$ determines three scheduling parameters: the TaylorSeer computation interval $\tau$, the Taylor expansion order $K$, and the action-chunk execution length $n_{\text{exec}}$.
We assign these parameters using heuristic stage-conditioned rules, based on the expected trade-offs between responsiveness, motion consistency, and inference efficiency in each stage:

In \emph{$\Phi_{\mathrm{static}}$}, we use a conservative schedule with frequent full recomputation and short execution chunks, since inaccurate early actions may lead to irreversible grasp failure.

In \emph{$\Phi_{\mathrm{dyn}}$}, tracking fast-moving targets demands minimal closed-loop latency. We adopt an aggressive Taylor~configuration with a large interval $\tau_{\text{follow}}$ and short action-chunk execution $n_{\text{short}}$, re-planning frequently so that corrections are applied at every cycle.

In \emph{$\Phi_{\mathrm{inter}}$}, maintaining contact quality and motion smoothness requires temporal coherence. We switch to a conservative setting with a smaller interval $\tau_{\text{interact}}$ and a longer action-chunk prefix $n_{\text{long}} > n_{\text{short}}$, prioritizing consistency over raw speed.

In \emph{$\Phi_{\mathrm{trans}}$},  returning has the least stringent timing requirements. We apply the most aggressive acceleration (largest interval, higher-order expansion) with full action-chunk execution, minimizing compute overhead during non-critical motions.

Overall, this heuristic stage-aware scheduler allocates computation according to the current interaction stage: it preserves frequent feedback in dynamic stages and reduces computation in less time-critical stages, thereby balancing responsiveness, motion consistency, and inference efficiency.
\begin{figure}[t]
  \centering
  \includegraphics[width=0.9\textwidth]{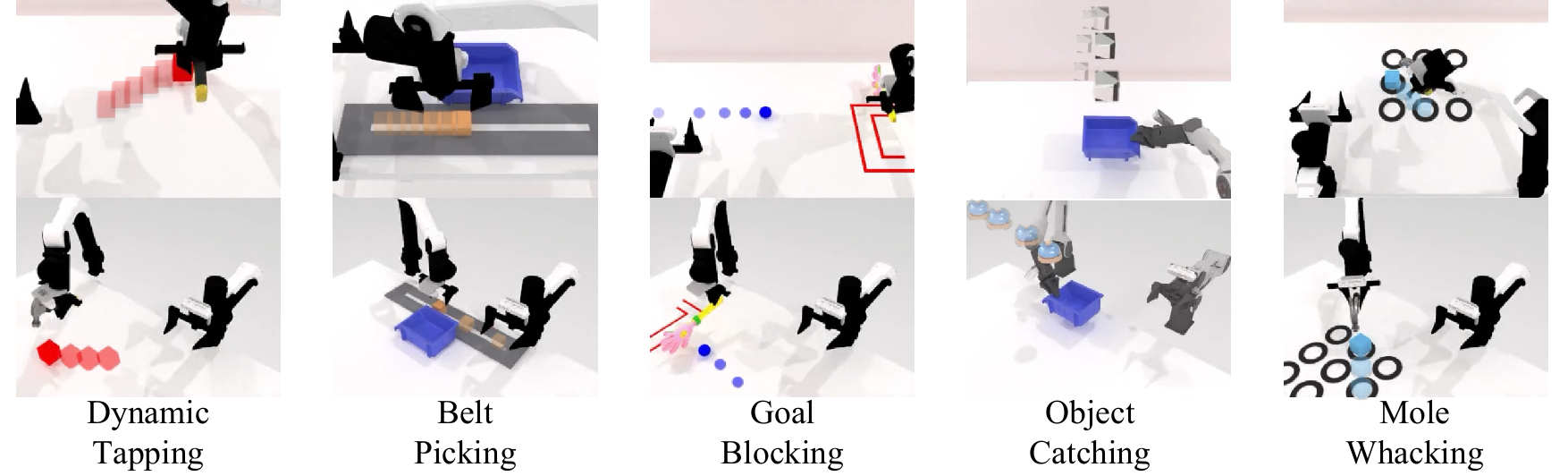}
  \vspace{-0.6em}
  \caption{Visualization of the tasks in the DynamicManip Simulation Benchmark.}
  \label{fig:benchmark_tasks}
  \vspace{-0.8em}
\end{figure}

\section{DynamicManip Benchmark}
\label{sec:benchmark}

To support the evaluation of dynamic manipulation capabilities, we introduce the DynamicManip Benchmark based on the RoboTwin 2.0 platform. We design five dynamic task environments for convenient data collection and policy evaluation, as illustrated in Figure~\ref{fig:benchmark_tasks}.

\textbf{Data Collection.} We extend the RoboTwin 2.0 pipeline with two key modifications tailored for dynamic manipulation. First, a dynamic-aware heuristic planner continuously tracks moving targets, predicts future positions, and triggers grasps at optimal moments. Second, a contact-release mechanism transfers control from scripted trajectories to the physics engine upon physical contact. These additions enable the robot to actually track and interact with moving objects, ensuring high-quality data collection. Finally, we record expert demonstrations capturing head-camera RGB, proprioceptive states, and scene step deltas at 30 FPS.

\textbf{Latency-Aware Automatic Evaluation.}
Our task-specific evaluator detects success using distinctive physical signatures (e.g., state-machine-based contact detection for tapping) and reports per-stage flags, trajectory residuals, and failure causes. Crucially, evaluation is coupled to the measured policy inference latency. Unlike conventional simulation, where inference time does not affect action execution, we advance the simulator by \(N_{\mathrm{delay}} = \operatorname{round}(t_{\mathrm{infer}} / \Delta t_{\mathrm{sim}})\) steps before applying each predicted action. The scene therefore evolves during inference, providing a harder and more realistic test of real-time responsiveness in dynamic tasks. Manual inspection of 100 episodes per task confirmed perfect agreement with human judgment.

\section{Experiments}
\label{sec:experiments}

We comprehensively evaluate the proposed DynamicManip framework in both simulated and real-world dynamic environments. Our experiments are designed to answer the following key questions:

\textbf{Q1}: Regarding data efficiency, \textbf{can our augmented data match or exceed human demonstrations} while significantly reducing the collection time?

\textbf{Q2}: In dynamic scenarios, \textbf{how does our dynamic-aware policy perform} in terms of reaction speeds and task success rates compared to the baseline?

\textbf{Q3}: \textbf{How do the key data-generation components and inference-scheduling variants affect task performance}?

\begin{figure}[t]
  \centering
  \includegraphics[width=0.9\textwidth]{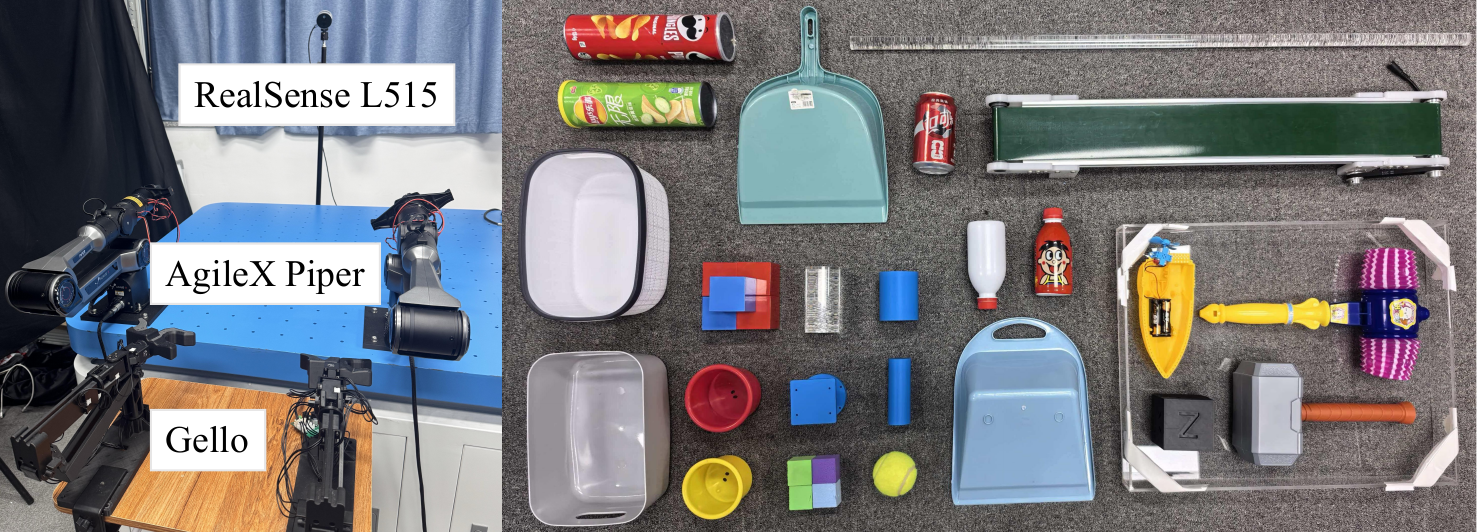}
  \vspace{-0.6em}
  \caption{Visualization of real-world robot setup and the objects used in our experiments.}
  \label{fig:realworld_setup}
  \vspace{-1.0em}
\end{figure}

\subsection{Experiment Settings}
\textbf{Simulation Setup.} We conduct simulation experiments to evaluate the performance of DynamicManip on the DynamicManip Benchmark. We use two ARX-X5 robotic arms with a pair of parallel grippers and a RealSense D435 head camera for visual perception. More details on the simulation environment can be found in the supplementary material.

\textbf{Real-World Setup.}
We also perform real-world experiments to validate the effectiveness of our approach.
We use a pair of AgileX Piper robots with parallel grippers and a RealSense L515 depth camera for visual perception.
As shown in Figure~\ref{fig:realworld_setup}, the experiments are conducted using objects with diverse geometries and physical properties.
The four real-world tasks and their autonomous policy inference and execution sequences are visualized in Appendix Figure~\ref{fig:realworld_tasks}.

\textbf{Evaluation Metrics.} We use the \emph{Success Rate} (Suc.) to measure the proportion of trials in which the robot successfully completes the dynamic task. Successful trials include cases such as hitting a moving target, catching an object, or intercepting a goal-bound item.

\textbf{Training and Inference Details.} 
The diffusion noise scheduler uses 100 timesteps during training. At inference, we use DDIM sampling with 10 denoising steps under our dynamic-aware computation schedule.
We train for 300 epochs with a batch size of 128. 
Optimization uses an initial learning rate of $2.0 \times 10^{-4}$, decayed via a cosine schedule. All experiments are implemented in PyTorch on a single RTX 4090 GPU for training, while real-world inference is performed on a single RTX 4070 GPU. Additional details are available in the supplementary material.

\textbf{Baseline.} We compare our method against the state-of-the-art imitation learning policy, 3D-Diffusion-Policy~\cite{ze20243dp3}, reproduced following the official RoboTwin defaults across every experiment, trained on the identical demonstration set, and evaluated under the same benchmark protocol, ensuring a fair comparison that isolates the contribution of our augmentation and inference components.

\begin{table}[t]
  \caption{Simulation data-efficiency comparison. All entries report success rate (\%). Upper rows use simulated demos; lower rows use dynamic episodes generated from one static
  demonstration.}
  \label{tab:data_efficiency}
  \vspace{-0.4em}
  
  \centering
  \setlength{\tabcolsep}{3.5pt}
  \begin{tabular}{llccccc}
    \toprule
    \multicolumn{2}{c}{\textbf{Setting}} & \textbf{\makecell{Dynamic\\Tapping}} & \textbf{\makecell{Belt\\Picking}} & \textbf{\makecell{Object\\Catching}} & \textbf{\makecell{Mole\\Whacking}}  & \textbf{\makecell{Goal\\Blocking}} \\
    \midrule
    \multicolumn{2}{l}{1 demo}            & $27$ & $0$ & $8$ & $3$ & $12$ \\
    \multicolumn{2}{l}{10 demos}          & $15$ & $5$ & $54$ & $21$ & $46$ \\
    \multicolumn{2}{l}{50 demos}          & $19$ & $31$ & $37$ & $63$ & $94$ \\
    \multicolumn{2}{l}{200 demos}         & $55$ & $85$ & $50$ & $83$ & $87$ \\
    \midrule
    \multicolumn{2}{l}{1 demo $\xrightarrow{\text{aug.}}$ 50 eps}  & $45$ & $44$  & $70$ & $63$ & $88$ \\
    \multicolumn{2}{l}{1 demo $\xrightarrow{\text{aug.}}$ 200 eps} & $52$ & $82$  & $64$ & $80$ & $88$ \\
    \bottomrule
  \end{tabular}
  \vspace{-0.8em}  
\end{table}

\subsection{Can Augmented Data Match or Exceed Human Demonstrations (Q1)}
\label{sec:data_scarcity}

To evaluate whether our augmentation pipeline can reduce the need for manually collected dynamic demonstrations, we compare policies trained on augmented data with those on conventional demonstrations.
We conduct this comparison using the DP3 baseline in both the simulated DynamicManip Benchmark and real-world setups, and report task performance and data-collection cost.


\textbf{Simulation analysis.}
A single static demonstration contains useful interaction structure, but training directly on it provides limited coverage of dynamic object motions and interaction timings.
To compare data sources and dataset scales under a consistent training protocol, we train the same DP3 policy configuration for 300 epochs across all settings, varying only the training data source and scale.
As shown in Table~\ref{tab:data_efficiency}, policies trained on our augmented episodes achieve success rates comparable to, and in some tasks higher than, policies trained on much larger sets of simulated demonstrations (e.g., 200 demos).
This result suggests that our augmentation pipeline can effectively expand one static demonstration into diverse dynamic training data. 
The gains come from combining geometric reconstruction with trajectory editing, which preserves the local robot-object interaction pattern while varying object motion, spatial placement, and phase timing.


\begin{table}[t]
  \centering
  \caption{Real-world data-efficiency comparison. We report success rates (\%) over 30 trials per task and setting. Dynamic Tapping contains three strikes per trial and is evaluated over 90 strike events; the other tasks use one binary outcome per trial. Higher is better.}
  \label{tab:realworld_data_efficiency}
  \setlength{\tabcolsep}{5pt}
  \renewcommand{\arraystretch}{1.05}
  \begin{tabular}{lcccc}
    \toprule
    \textbf{Training Data}
    & \makecell{\textbf{Dynamic}\\\textbf{Tapping}}
    & \makecell{\textbf{Belt}\\\textbf{Picking}}
    & \makecell{\textbf{Boat}\\\textbf{Loading}}
    & \makecell{\textbf{Bottle}\\\textbf{Catching}} \\
    \midrule
    Manual dyn. demos (1 demo)
    & $6.67$ & $30.0$ & $0.00$ & $0.00$ \\

    Manual dyn. demos (50 demos)
    & $68.9$ & $70.0$ & $3.33$ & $73.3$ \\
    \midrule
    Ours: 1 static demo $\rightarrow$ 50 aug. eps
    & $61.1$ & $56.7$ & $23.3$ & $76.7$ \\

    Ours: 1 static demo $\rightarrow$ 200 aug. eps
    & $\mathbf{87.8}$ & $\mathbf{73.3}$ & $\mathbf{66.7}$ & $\mathbf{83.3}$ \\
    \bottomrule
  \end{tabular}
\end{table}

\textbf{Real-world validation}
We further evaluate whether the augmented data remain effective in real-world physical setups (Table~\ref{tab:realworld_data_efficiency}).
Starting from a single static demonstration, increasing the number of augmented episodes consistently improves policy performance.
With sufficient augmented data, the policy surpasses the success rate of a model trained on 50 manually teleoperated dynamic demonstrations.
This result indicates that the proposed augmentation pipeline can transfer the benefit of static-to-dynamic data generation to real-world dynamic manipulation tasks.

\begin{table}[t]
  \centering
  \caption{Human operators and elapsed human/machine time for data preparation. Our pipeline is separated into static teleoperation, pre-generation processing, and automated trajectory generation; Appendix Table~\ref{tab:task_setup_cost} provides the complete processing breakdown.}
  \label{tab:time_cost}
  \renewcommand{\arraystretch}{1.08}
  {\small
  \setlength{\tabcolsep}{3pt}
  \begin{tabular}{@{}lccc@{}}
    \toprule
    \textbf{Data Preparation Setting}
    & \textbf{Human Operators}
    & \textbf{Human Time}
    & \textbf{Machine Time (RTX 4090)} \\
    \midrule
    Manual teleop. (50 dyn. demos)
    & $2$ & $\sim100$ min & -- \\

    Manual teleop. (200 dyn. demos)
    & $2$ & $\sim400$ min & -- \\
    \midrule
    Ours: teleop. (1 static demo)
    & $1$ & $\sim30$ s & -- \\

    Ours: pre-generation processing
    & $1$ & $\sim10$ min & $\sim1$ min \\

    Ours: automated aug. (50 eps)
    & $0$ & -- & $\sim7$ min \\

    Ours: automated aug. (200 eps)
    & $0$ & -- & $\sim30$ min \\
    \midrule
    \textbf{Ours: total (200 eps)}
    & $\mathbf{1}$ & $\mathbf{\sim11}$ min & $\mathbf{\sim31}$ min \\
    \bottomrule
  \end{tabular}
  }
  \vspace{-0.8em}
\end{table}

\textbf{Human effort and collection time.} Table~\ref{tab:time_cost} compares the human effort and machine time required for data preparation. Manual dynamic demonstrations require sustained synchronized teleoperation, whereas our pipeline uses only a short static demonstration and limited task-specific processing before automated trajectory generation. This substantially reduces human involvement while shifting the remaining data-generation cost to machine computation; Appendix Table~\ref{tab:task_setup_cost} provides the complete processing breakdown.

\textbf{Summary.}
These results strongly support an affirmative answer to \textbf{Q1}: DynamicManip can synthesize highly effective dynamic training data from a single static demonstration.
Across simulation and real-world experiments, policies trained on our augmented data match the performance of policies trained on substantially more conventional demonstrations and exceed it in several simulation tasks and all evaluated real-world tasks.
This effectiveness comes from preserving local robot-object interaction patterns while expanding the coverage of object motion, spatial placement, and interaction timing,
which helps the policy learn the underlying dynamics of object motion and robot-object interaction.
Crucially, DynamicManip achieves this performance with lower manual data collection cost, making dynamic manipulation learning more data-efficient and practical in real-world settings.

\begin{table}[t]
  \caption{Latency and success-rate comparison between DP3 and our
  dynamic-aware policy under the Latency-Aware Automatic Evaluation System. Lat. means end-to-end inference latency measured in milliseconds(ms), success
  rate is evaluated under 100 trials and reported as a percentage(\%).}
  \label{tab:baseline_comparison}
  \centering
  \setlength{\tabcolsep}{3.5pt}
  \begin{tabular}{l *{5}{cc}}
    \toprule
    & \multicolumn{2}{c}{\textbf{\makecell{Dynamic\\Tapping}}}
    & \multicolumn{2}{c}{\textbf{\makecell{Belt\\Picking}}}
    & \multicolumn{2}{c}{\textbf{\makecell{Object\\Catching}}}
    & \multicolumn{2}{c}{\textbf{\makecell{Mole\\Whacking}}}
    & \multicolumn{2}{c}{\textbf{\makecell{Goal\\Blocking}}} \\
    \textbf{Method}
    & Lat. & Suc.
    & Lat. & Suc.
    & Lat. & Suc.
    & Lat. & Suc.
    & Lat. & Suc. \\
    \midrule
    DP3~\cite{ze20243dp3}
    & $60.0$ & $20.0$
    & $45.0$ & $78.0$
    & $61.4$ & $48.0$
    & $44.4$ & $38.0$
    & $44.6$ & $42.0$ \\
    \midrule
    \textbf{Ours}
    & $\mathbf{38.0}$ & $\mathbf{42.0}$
    & $\mathbf{32.9}$ & $\mathbf{80.0}$
    & $\mathbf{36.0}$ & $\mathbf{64.0}$
    & $\mathbf{30.6}$ & $\mathbf{44.0}$
    & $\mathbf{33.9}$ & $\mathbf{88.0}$ \\
    \bottomrule
  \end{tabular}
  \vspace{-0.8em}
\end{table}

\begin{table*}[t]
  \caption{Transposed ablation of cumulative data-generation variants and inference schedules (success rate, \%). Scheduling variants use the same checkpoint and evaluation episodes. Ours additionally includes local interaction replay and predicted-stage scheduling.}
  \label{tab:ablation}
  \centering
  \small
  \renewcommand{\arraystretch}{0.9}
  \setlength{\tabcolsep}{5pt}

  \begin{tabular*}{\textwidth}{@{\extracolsep{\fill}}lccccccc@{}}
    \toprule
    \multirow{2}{*}{\textbf{Task}}
    & \multicolumn{3}{c}{\textbf{Cumulative Data Generation}}
    & \multicolumn{3}{c}{\textbf{Inference Scheduling}}
    & \multirow{2}{*}{\textbf{Ours}} \\
    \cmidrule(lr){2-4}\cmidrule(lr){5-7}
    & \makecell{Spatial\\Only}
    & \makecell{Quasi-\\static}
    & \makecell{Planned\\Motion}
    & \makecell{Uniform\\DDIM}
    & \makecell{Fixed\\Taylor}
    & \makecell{Short\\Chunks}
    & \\
    \midrule
    Object Catching
    & $2$ & $26$ & $53$
    & $19$ & $34$ & $48$
    & $\mathbf{64}$ \\

    Goal Blocking
    & $4$ & $34$ & $75$
    & $38$ & $67$ & $64$
    & $\mathbf{88}$ \\
    \bottomrule
  \end{tabular*}
  \vspace{-0.8em}
\end{table*}
\subsection{Real-Time Performance of the Dynamic-Aware Policy (Q2)}
\label{sec:realtime_response}
To isolate the effect of the dynamic-aware policy, we compare DynamicManip with DP3 using the same demonstration set and latency-aware evaluation protocol. Table~\ref{tab:baseline_comparison} reports end-to-end policy-query latency and success rate over 100 trials per task.

Across the five tasks, DynamicManip reduces mean policy-query latency by 32.9\% and increases mean success by 18.4 points. It improves both metrics on every task, with the largest success gain on Goal Blocking (+46 percentage points). These results show that stage-aware adaptive inference improves real-time responsiveness without trading off task success under latency-aware execution.

\subsection{Ablation Studies on Data Generation and Inference Scheduling (Q3)}
\label{sec:ablation}

Table~\ref{tab:ablation} reports two controlled ablations. For data generation, we cumulatively add spatial transforms, quasi-static editing, planned dynamic motion, and local interaction replay while keeping the policy architecture, training configuration, and evaluation protocol fixed. For inference scheduling, we use the same checkpoint and evaluation episodes and change only the scheduling strategy.

Each successive addition improves both tasks, indicating that the data-generation components are complementary rather than redundant. Planned dynamic motion provides the largest stepwise gain, while local interaction replay yields a further improvement. Under the controlled scheduling comparison, our predicted-stage strategy performs best on both tasks, indicating that jointly adapting denoising, Taylor acceleration, and action-chunk length to the predicted stage is more effective than using a fixed schedule under this evaluation.

\section{Conclusion}
\label{sec:conclusion}

In this paper, we present \textit{DynamicManip} to address the critical bottlenecks of data scarcity and inference latency in dynamic robot manipulation. 
We introduce an efficient static-to-dynamic augmentation pipeline that synthesizes diverse, physically plausible demonstrations from only a single static demonstration, alongside a dynamic-aware adaptive policy that automatically regulates inference frequency for responsive closed-loop control. 
Furthermore, we establish a comprehensive dynamic manipulation benchmark with an automatic evaluation system to provide a scalable assessment standard. 
Extensive experiments in both simulation and real-world settings demonstrate that our approach significantly improves data efficiency and task success rates while achieving faster inference than existing baselines. 
By effectively bridging the gap between minimal static data and complex reactive execution, our framework represents a robust and scalable step toward deploying robots in unstructured, fast-paced human environments.

\newpage

\bibliographystyle{plainnat}
\bibliography{references}

\newpage

\appendix
\section*{Appendix}
\renewcommand{\thesection}{\Alph{section}}

\section{Network Architecture \& Training Details}
\label{app:architecture}

\subsection{Network Architecture}

Our model adopts a diffusion-based framework that incorporates dynamic-aware perception and action generation capabilities. The architecture is designed to process multi-modal observations and generate precise robot actions through a denoising diffusion process.

\subsubsection{Input modalities.} The model takes two types of inputs at each timestep:

\texttt{Point clouds:} Single-view point clouds serve as high-dimensional visual observations, denoted as $\mathbf{o}_t^h \in \mathbb{R}^{N \times 3}$, where $N = 2048$ points are obtained via Farthest Point Sampling (FPS). We define the observation horizon as $t_o$ and in our experiments, we set $t_o=6$. 

\texttt{Proprioception:} Low-dimensional robot state observations $\mathbf{o}_t^l \in \mathbb{R}^{M}$ capture joint positions and gripper configurations across a temporal window. The proprioception shares the observation horizon as $t_o$.






\paragraph{Point Cloud Encoder.} We first crop the point cloud using a 3D bounding box and downsample to $N = 2048$ points via FPS. We employ a PointNet~\cite{charles2017pointnet} encoder to independently encode each of the $K_{\text{high}}$ sampled frames, then concatenate the features along the channel dimension to obtain the final visual representation $\mathbf{f}^h$.

\paragraph{State Encoder.} Proprioceptive observations are encoded \emph{per frame} with a lightweight MLP. For each observation step, the robot state vector is projected into a state feature token via the MLP. The state feature is then concatenated with the point-cloud feature at the same timestep.

\subsubsection{Loss Function}

We employ FiLM \cite{perez2018film} conditioning to inject the conditional feature $f_c$ into the denoising network, which predicts the noise associated with the robot action $\mathbf{a} \in \mathbb{R}^{k_a}$, where $k_a$ denotes the action dimension specific to the task. The primary training objective minimizes the denoising score matching loss:
\begin{equation}
\mathcal{L}_{\text{diff}} = \mathbb{E}_{\mathbf{a}_0, \boldsymbol{\epsilon} \sim \mathcal{N}(0, \mathbf{I})} \left[ \left\| \boldsymbol{\epsilon} - \epsilon_\theta(\mathbf{a}_t, t) \right\|^2 \right],
\end{equation}
where $\boldsymbol{\epsilon} \sim \mathcal{N}(0, \mathbf{I})$ is the Gaussian noise, and the network $\epsilon_\theta$ is trained to predict the added noise given the noisy action $\mathbf{a}_t$ and the diffusion timestep $t$. We employ DDIM \cite{song2020ddim} for inference sampling.

\textbf{Dynamic-Stage Awareness Learning} Every augmented trajectory carries a stage label $\ell_t \in \{\Phi_{\mathrm{static}}, \Phi_{\mathrm{dyn}}, \Phi_{\mathrm{inter}}, \Phi_{\mathrm{trans}}\}$ at each timestep $t$, reflecting the interaction logic that produced it. We leverage these labels as free supervision signals by attaching a lightweight classification head $h_\phi$ to the shared observation encoder. Given the fused feature $f_{lh}$ processed through a dense fusion network $\phi_{\text{dense}}$, the head predicts a stage distribution $\hat{\mathbf{p}}_t = h_\phi(\phi_{\text{dense}}(f_{lh}))$ over the four stages, supervised by a cross-entropy loss:
\begin{equation}
\mathcal{L}_{\text{stage}} = -\frac{1}{T}\sum_{t=1}^{T} \log \hat{p}_t(\ell_t).
\end{equation}
The dense fusion network $\phi_{\text{dense}}$ enhances feature representations that are subsequently used by the diffusion decoder, while $h_\phi$ is intentionally kept as a lightweight single-layer MLP to mitigate the risk of overfitting the auxiliary head. The total training objective combines both losses:
\begin{equation}
\mathcal{L} = \mathcal{L}_{\text{diff}} + \lambda \mathcal{L}_{\text{stage}},
\end{equation}
where $\lambda = 0.1$ balances action fidelity and stage awareness. This auxiliary objective encourages the encoder to learn representations that disambiguate interaction phases, which are later exploited at inference time with negligible additional computation.

\subsection{Training Configuration}

We train all models using the AdamW optimizer with an initial learning rate of $2.0 \times 10^{-4}$, betas $[0.95, 0.999]$, weight decay $1 \times 10^{-6}$, and batch size 128. Training proceeds for 300 epochs with a cosine learning rate schedule and 500 warmup steps. We employ Exponential Moving Average (EMA) with momentum 0.9999 on model parameters, activated after the first training step.

The noise scheduler follows a squared cosine schedule with 100 diffusion timesteps during training, with $\beta_{\text{start}}=0.0001$ and $\beta_{\text{end}}=0.02$. For inference, we use DDIM sampling with 10 denoising steps to efficiently generate action sequences.

Data augmentation is minimal to preserve the precise geometric relationships required for manipulation: we only apply random translation within $\pm$2cm in the horizontal plane during point cloud preprocessing. All models are trained on a single NVIDIA RTX 4090 GPU.

\subsection{Inference}
\label{sec:appendix:inference}

\subsubsection{Inference Sampling.} We employ DDIM for efficient inference sampling. The reverse diffusion process is formulated as:
\begin{equation}
    \mathbf{a}_{t-1} = \sqrt{\bar{\alpha}_{t-1}} \left( \frac{\mathbf{a}_t - \sqrt{1 - \bar{\alpha}_t} \epsilon_\theta(\mathbf{a}_t, t)}{\sqrt{\bar{\alpha}_t}} \right) + \sqrt{1 - \bar{\alpha}_{t-1}} \epsilon_\theta(\mathbf{a}_t, t),
\end{equation}
where $\bar{\alpha}_{t-1}$ and $\bar{\alpha}_t$ are the cumulative noise schedule coefficients at timesteps $t-1$ and $t$, respectively. We use 10 denoising steps during inference for efficient real-time deployment.

\subsubsection{TaylorSeer Acceleration}

During inference, the diffusion model must solve the reverse denoising process iteratively, which requires multiple forward passes through the 1D U-Net. Each DDIM step involves computing activations for all residual blocks across three resolution levels, making inference computationally expensive and limiting the real-time control frequency of the policy. To address this bottleneck, we introduce a Taylor-based acceleration mechanism that replaces selective U-Net forward passes with Taylor series extrapolation during the denoising trajectory.

\textbf{Principle.} The core observation is that adjacent DDIM steps produce highly correlated intermediate features --- the U-Net's residual block outputs vary smoothly as the noise level changes incrementally along the reverse trajectory. Let $f^{(l)}(t_i)$ denote the output of residual block $l$ at denoising step $t_i$. We maintain a cache of Taylor coefficients computed via backward finite differences over consecutive steps:
\begin{equation}
    c_0 = f^{(l)}(t_i), \quad
    c_1 = \frac{f^{(l)}(t_i) - f^{(l)}(t_{i-1})}{t_i - t_{i-1}}, \quad
    c_2 = \frac{c_1^{(i)} - c_1^{(i-1)}}{t_i - t_{i-1}},
\end{equation}
and so on up to the maximum order $K$. For a future step $t_{i+\delta}$ where we skip the full forward pass, the Taylor prediction is:
\begin{equation}
    \hat{f}^{(l)}(t_{i+\delta}) = \sum_{j=0}^{K} \frac{c_j}{j!} \, (t_{i+\delta} - t_i)^j.
\end{equation}

\textbf{Effect.}
TaylorSeer reduces inference latency by avoiding redundant full U-Net evaluations at selected DDIM steps.
The latency values reported in the main paper are measured as end-to-end policy-query latency per control cycle, from the arrival of an observation to the output of the action chunk.
This measurement includes observation encoding, DDIM sampling, stage-aware scheduling, and action selection.

For clarity, we use module-level profiling only to characterize the denoising computation itself.
On a single NVIDIA RTX 4090 GPU, replacing a selected full U-Net computation with Taylor extrapolation reduces the cost of that denoising computation from approximately $46$ ms to $22$ ms, corresponding to about a $2.09\times$ speedup for the approximated computation.
The actual end-to-end latency depends on the stage-specific schedule, the number of full recomputation steps, and non-denoising overheads.
We therefore report the measured end-to-end latency in Table~\ref{tab:baseline_comparison} of the main paper rather than deriving it by analytically multiplying the per-step profiling numbers.
This acceleration is especially beneficial for real-time robot control, where the policy must repeatedly sample action trajectories under a strict control-frequency budget. Since Taylor operates only on intermediate U-Net features during inference, it does not modify the diffusion training objective, the DDIM update rule, or the policy output representation, making it a plug-and-play acceleration module for the deployed diffusion policy.

\textbf{Computation Scheduling}
We do not apply Taylor approximation at every step. Instead, we adopt a periodic recomputation strategy that interleaves full forward passes with Taylor predictions. Specifically:
\begin{itemize}
    \item \textbf{First step:} always computed fully to establish the initial anchor.
    \item \textbf{First $k$ steps (default $k=3$):} always computed fully, as the early denoising steps contain the most informative signal about the target action distribution.
    \item \textbf{Every $\tau$ steps (default $\tau=5$):} computed fully to refresh the Taylor coefficients and prevent extrapolation drift.
    \item \textbf{All other steps:} replaced by Taylor prediction of order $K$ (default $K=1$, i.e., linear extrapolation).
\end{itemize}

With $\tau=5$, $K=1$, and $k=3$ under $T=10$ DDIM inference steps, 6 out of 10 steps use Taylor prediction while only 4 steps require full U-Net evaluation, reducing the number of full forward passes by approximately $60\%$.

\textbf{Per-Stage Adaptive Scheduling}
Different interaction stages have distinct requirements for inference accuracy and re-planning frequency. We define four task stages corresponding to the data augmentation phases: $\Phi_{\mathrm{static}}$ (static object acquisition), $\Phi_{\mathrm{dyn}}$ (dynamic target alignment), $\Phi_{\mathrm{inter}}$ (contact interaction execution), and $\Phi_{\mathrm{trans}}$ (pose-conditioned transition). For each stage, the predicted stage classifier output $\hat{\ell}_t$ determines three scheduling parameters: the TaylorSeer computation interval $\tau$, the Taylor expansion order $K$, and the number of executed action steps $n_{\text{exec}}$ from the predicted action chunk. The scheduling parameters are listed in Table~\ref{tab:stage_scheduling}.

\begin{table}[h]
\centering
\small
\begin{tabular}{@{}lccc@{}}
\toprule
\textbf{Stage} & $\boldsymbol{\tau}$ (interval) & $\boldsymbol{K}$ (order) & $\boldsymbol{n_{\text{exec}}}$ \\
\midrule
$\Phi_{\mathrm{static}}$  & 3 & 1 & 2 \\
$\Phi_{\mathrm{dyn}}$     & 7 & 1 & 2 \\
$\Phi_{\mathrm{inter}}$   & 3 & 1 & 8 \\
$\Phi_{\mathrm{trans}}$   & 8 & 2 & 8 \\
\bottomrule
\\
\end{tabular}
\caption{Stage-specific heuristic TaylorSeer and execution step scheduling. $\tau$ is the recomputation interval, $K$ is the Taylor polynomial order, and $n_{\text{exec}}$ is the number of action steps executed per inference call.}
\label{tab:stage_scheduling}
\end{table}

\subsubsection{Execution Step Configuration}

The policy operates in a receding horizon control loop. At each inference call, the model predicts an action chunk of length $H=8$ (the \textit{action horizon}) via DDIM sampling with $T=10$ denoising steps. Only a subset of $n_{\text{exec}}$ actions from this chunk are actually executed before the next inference call is triggered. The remaining actions are discarded, and a fresh prediction is computed from the updated observations.

In our framework, $n_{\text{exec}}$ is adjusted per Table~\ref{tab:stage_scheduling}. 
The static acquisition stage uses frequent recomputation and short execution chunks to avoid early grasp errors.
The dynamic alignment stage keeps short chunks for frequent feedback while using a larger Taylor interval to reduce latency.
The interaction and transition stages execute longer chunks because they benefit more from motion continuity and are less dependent on immediate visual correction.

\section{Efficient Data Augmentation Pipeline}
\label{appendix:augmentation_pipeline}

In this section, we provide a detailed description of the data augmentation pipeline used in our framework. The process begins with a single \textbf{static} demonstration, from which we systematically synthesize massive and diverse augmented episodes featuring complex dynamic target movements. To seamlessly scale up the training dataset, our augmentation engine is parallelized across a cluster of four NVIDIA RTX 4090 GPUs. This fully automated pipeline encompasses several critical computational steps—including mesh generation, precise mesh alignment, dynamic trajectory editing, and point cloud post-processing—which we describe in detail below.

\subsection{Task-Specific Setup Cost}
\label{app:task_setup_cost}

Task-specific data preparation requires only limited human intervention.
The planner, reachability constraints, and motion primitives are reused
across tasks, while only a small number of task-specific components need
to be configured.

\begin{table}[t]
  \centering
  \caption{Task-specific setup and data-generation time.
  All times are approximate and reported in minutes.}
  \label{tab:task_setup_cost}

  \small
  \renewcommand{\arraystretch}{1.08}
  \setlength{\tabcolsep}{6pt}

  \begin{tabular}{@{}p{0.62\linewidth}cc@{}}
    \toprule
    \textbf{Step}
    & \textbf{Human}
    & \textbf{Machine} \\
    \midrule

    One static teleoperated demonstration
    & 1 & -- \\

    Keyframe annotation
    & 2 & 1 \\

    Mesh-size correction
    & 2 & -- \\

    Phase composition
    & 1 & -- \\

    Motion-distribution specification
    & 5 & -- \\

    Automated generation of 200 trajectories
    & -- & 30 \\

    \midrule
    \textbf{Total}
    & \textbf{11}
    & \textbf{31} \\

    \bottomrule
  \end{tabular}

  \vspace{0.4em}

  \parbox{0.96\linewidth}{%
    \footnotesize
    \textit{Note.}
    When keyframes are annotated online during demonstration collection,
    the total cost is reduced to approximately 9 minutes of human effort
    and 30 minutes of machine computation.
    The planner, reachability constraints, and motion primitives are reused
    across tasks.
  }
\end{table}

\subsection{Mesh Generation}

To ensure the precise geometric reconstruction of the manipulation scene, we capture an RGB image of the manipulation object and generate a 3D mesh using a 3D generative model~\cite{hyper3d2024}. This generated mesh is then utilized in the subsequent mesh alignment step, where it is integrated to form a complete 3D representation of the object. This step is essential for accurate trajectory editing and the generation of realistic augmented episodes.

\subsection{Mesh Alignment and Geometric Reconstruction}

A crucial part of our data augmentation pipeline is the alignment of known CAD meshes to the observed point cloud. Since point clouds captured from a single viewpoint typically only contain visible surfaces, this step helps in reconstructing complete 3D assets, which are then used in subsequent stages.

\subsubsection{Coarse-to-Fine Mesh Alignment}

The mesh alignment is performed using a coarse-to-fine approach, where an initial alignment is followed by a refinement step to improve accuracy. The steps are as follows:

\begin{itemize}
    \item \textbf{Coarse Alignment (OBB-based Initialization):} We begin by aligning the mesh to the observed point cloud using an Oriented Bounding Box (OBB). The OBB is computed to estimate the initial transformation between the mesh and the point cloud.
    \item \textbf{Fine Alignment (Iterative Closest Point - ICP):} After the coarse alignment, we apply a multi-resolution ICP algorithm to iteratively refine the alignment. ICP minimizes the distance between the points in the point cloud and the vertices of the mesh. This is achieved by solving for the transformation that best matches the observed points to the mesh at different resolutions.
\end{itemize}

The mathematical formulation of the ICP step is as follows:
\[
    \min_{\mathbf{T}} \sum_{i=1}^{N} \left\| \mathbf{p_i} - \mathbf{T} \mathbf{m_i} \right\|^2
\]
where \( \mathbf{p_i} \) is the point from the observed point cloud, \( \mathbf{m_i} \) is the corresponding mesh vertex, and \( \mathbf{T} \) is the transformation matrix that minimizes the point-to-point distance.

\begin{figure}
    \centering
    \includegraphics[width=1\textwidth]{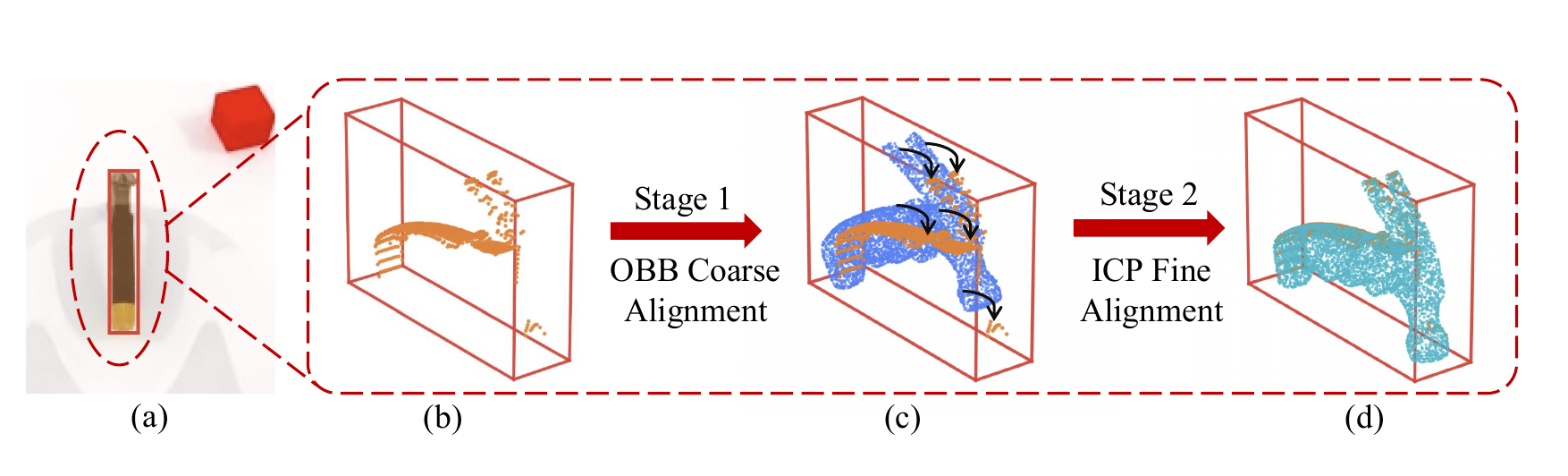}
    \caption{Visualization of the mesh alignment process. (a) shows the ego view image of a manipulation object, (b) shows the initial point cloud captured from the RGB-D sensor, (c) shows the coarse alignment using OBB, and (d) shows the refined alignment after applying ICP. The aligned mesh is then used for geometric reconstruction in the augmentation pipeline.}
    \label{fig:mesh_alignment}
\end{figure}

The visualization of the mesh alignment process is shown in Figure~\ref{fig:mesh_alignment}.

\subsubsection{Canonical Point Cloud Generation}

Once the mesh is aligned, the aligned mesh is resampled into a canonical point cloud that is spatially invariant. This canonical point cloud is used in subsequent steps and allows for geometrically coherent transformations during editing.

\subsection{Dynamic Trajectory Editing}
\label{subsec:appendix_trajectory_editing}

The core of our data augmentation pipeline lies in the \textit{Dynamic Trajectory Editing} engine, which systematically decomposes a single static demonstration into four distinct physical phases: Static Object Acquisition ($\Phi_{\mathrm{static}}$), Dynamic Target Alignment ($\Phi_{\mathrm{dyn}}$), Contact Interaction Execution ($\Phi_{\mathrm{inter}}$), and Pose-Conditioned Transition ($\Phi_{\mathrm{trans}}$). Rather than relying on simple linear interpolations, our engine dynamically synthesizes and warps trajectories by coupling geometric registration, physical priors, and real-time motion planning. This ensures that the generated trajectories are not only physically feasible but also exhibit robust spatial generalization.

\subsubsection{Static Object Acquisition ($\Phi_{\mathrm{static}}$)}

In tasks involving auxiliary tools (e.g., grasping a hammer or picking up a basket), the initial segment of the demonstration involves grasping or acquiring a static object. Because the object remains stationary during this phase, we parameterize this segment by annotating its start and end keyframes, denoted as $t_{\mathrm{start}}^{\mathrm{static}}$ and $t_{\mathrm{end}}^{\mathrm{static}}$, respectively. 

To generalize this acquisition behavior to arbitrary initial poses in the augmented environment, we apply a rigid transformation $T_k^{\mathrm{static}} \in \mathbb{SE}(3)$ to the recorded trajectory. The modified end-effector pose $\mathbf{e}'_t$ and object trajectory $\mathbf{p}'_t$ at any time step $t \in [t_{\mathrm{start}}^{\mathrm{static}}, t_{\mathrm{end}}^{\mathrm{static}}]$ are generated by replaying the source trajectory under this transformation:
\begin{equation}
    \mathbf{p}'_t = T_k^{\mathrm{static}} \mathbf{p}_t, \quad \mathbf{e}'_t = T_k^{\mathrm{static}} \mathbf{e}_t
\end{equation}
This rigid formulation effortlessly ensures spatial generalization, allowing the policy to reliably acquire tools or objects under diverse spatial displacements without requiring additional manual demonstrations.

\subsubsection{Dynamic Target Alignment ($\Phi_{\mathrm{dyn}}$)}

This phase aligns the end-effector with a moving object or a predicted interaction target before contact.
Instead of collecting dynamic teleoperation for this alignment behavior, we compute a time-varying target from the synthesized object motion and a task-specific prediction rule.
Formally, given the synthesized object trajectory $T_{\mathrm{obj}}^{k}(t)$ and augmentation parameters $\xi_k$, the interaction target is generated as
\begin{equation}
    \mathbf{g}^{k}(t) = \Gamma_{\mathrm{task}}\!\left(T_{\mathrm{obj}}^{k}(0:t), \xi_k\right),
\end{equation}
where $\Gamma_{\mathrm{task}}$ denotes the task-specific target-generation rule.
For object-relative interactions, $\mathbf{g}^{k}(t)$ can be an end-effector pose computed from a relative transform extracted from a contact keyframe:
\begin{equation}
    \mathbf{g}^{k}(t) = T_{\mathrm{obj}}^{k}(t) T_{\mathrm{offset}} T_{\mathrm{rel}}^{*}.
\end{equation}
For interception or catching tasks, $\mathbf{g}^{k}(t)$ can instead be a predicted interception or landing target computed from the object's motion model, such as a parabolic trajectory estimate.

To reach this dynamic target, a reactive motion planner generates smooth trajectories under a maximum velocity constraint $v_{\max}$.
The planner replans at a high frequency to correct tracking deviations and projects the resulting trajectory into the robot's reachable workspace, cropping segments that violate kinematic limits.
The alignment phase terminates once the end-effector is sufficiently close to the generated target under a task-defined distance metric $d(\cdot,\cdot)$:
\begin{equation}
    d\!\left(\mathbf{e}(t), \mathbf{g}^{k}(t)\right) < \epsilon.
\end{equation}
This formulation matches the main paper's generalized target-conditioned alignment phase while still retaining fixed object-frame relative poses as a special case.

\subsubsection{Contact Interaction Execution ($\Phi_{\mathrm{inter}}$)}

The contact phase captures the precise, high-frequency interaction behaviors between the robot and the object, such as striking a target or executing a dynamic tap. Since these local contact patterns are highly sensitive to physical constraints and difficult to plan from scratch, we preserve the local kinematic interaction pattern.

This phase is bounded by annotated start and end keyframes $[t_{\mathrm{start}}^{\mathrm{inter}}, t_{\mathrm{end}}^{\mathrm{inter}}]$. We model the local trajectory relative to a source anchor frame $A$, which represents the contact state in the demonstration. During augmentation, we warp this segment to a new target anchor $A_k$ (determined by the terminal pose of the preceding $\Phi_{\mathrm{dyn}}$ phase), translating and rotating the entire interaction sequence:
\begin{equation}
    \mathbf{e}'_t = A_k A^{-1} \mathbf{e}_t, \quad \forall t \in [t_{\mathrm{start}}^{\mathrm{inter}}, t_{\mathrm{end}}^{\mathrm{inter}}]
\end{equation}
By shifting the anchor frame, our engine replays the high-fidelity contact behavior at arbitrary spatial coordinates, seamlessly matching the physical constraints of different task locations while preserving the local kinematic interaction pattern.

\subsubsection{Pose-Conditioned Transition ($\Phi_{\mathrm{trans}}$)}

After completing an interaction, the robot must safely transition to a target spatial configuration, such as returning to a neutral home position or moving to an intermediate waypoint (e.g., returning to the center in a whack-a-mole task). 

Rather than replaying a rigid, collision-prone path, we leverage our motion planner to generate collision-free trajectories to a transformed keyframe pose. This phase requires only a single keyframe annotation to identify the target pose $\mathbf{e}_{\mathrm{key}}$. To introduce spatial variation and prevent policy overfitting, we apply a stochastic pose offset $T_k^{\mathrm{trans}} \in \mathbb{SE}(3)$ to the target configuration:
\begin{equation}
    \mathbf{e}_{\mathrm{key}}^{k} = T_k^{\mathrm{trans}} \mathbf{e}_{\mathrm{key}}
\end{equation}
The motion planner then computes a smooth trajectory from the robot's current state to $\mathbf{e}_{\mathrm{key}}^{k}$. The integration of the motion planner ensures that the robot can robustly transition across diverse configurations while satisfying collision and joint-limit constraints.

\subsection{Task-Specific Phase Composition Logic}
\label{appendix:task_composition}

Unlike monolithic trajectory generation paradigms that frequently suffer from compounding errors during long-horizon execution, the core strength of \textit{DynamicManip} lies in its ability to decouple complex manipulation into a precise sequence of modular phase operators $\{\Phi_{\mathrm{static}}, \Phi_{\mathrm{dyn}}, \Phi_{\mathrm{inter}}, \Phi_{\mathrm{trans}}\}$. By strategically recomposing these operators, our framework effortlessly instantiates diverse, long-horizon interaction patterns while maintaining strict physical plausibility. This section details the precise composition formulas and the underlying tracking logic used across both simulation and real-world benchmarks. geometric and kinematic consistency under our modeling assumptions

\subsubsection{Simulation Benchmark Tasks}

The interaction logic for simulation tasks is designed to robustly handle both infinite cyclic behaviors and high-precision spatial interceptions, ensuring zero-shot generalization to varying target velocities.

\begin{itemize}
    \item \textbf{Dynamic Tapping:} To synthesize continuous striking motions without accumulating temporal drift, we employ a tightly coupled cyclic structure: 
    $\Xi_{\text{tap}} = \Phi_{\mathrm{static}} \rightarrow (\Phi_{\mathrm{dyn}} \rightarrow \Phi_{\mathrm{inter}})^L \rightarrow \Phi_{\mathrm{trans}}$. 
    The policy first stably acquires the tool ($\Phi_{\mathrm{static}}$). Subsequently, the motion planner continuously guarantees real-time target alignment ($\Phi_{\mathrm{dyn}}$) before strictly executing the high-frequency striking template ($\Phi_{\mathrm{inter}}$). The final $\Phi_{\mathrm{trans}}$ ensures a safe return to the reset configuration.
    
    \item \textbf{Mole Whacking:} Because the target's reappearance location is highly unpredictable, this task requires a full spatial reset after every interaction to prevent out-of-distribution tracking angles: 
    $\Xi_{\text{mole}} = \Phi_{\mathrm{static}} \rightarrow (\Phi_{\mathrm{dyn}} \rightarrow \Phi_{\mathrm{inter}} \rightarrow \Phi_{\mathrm{trans}})^L$. 
    By forcing a $\Phi_{\mathrm{trans}}$ step within the loop, the robot consistently restores an optimal central vantage point.
    
    \item \textbf{Object Catching:} To maximize dynamic responsiveness, this task strips away non-essential interactions: 
    $\Xi_{\text{catch}} = \Phi_{\mathrm{static}} \rightarrow \Phi_{\mathrm{dyn}}$. 
    The robot strictly focuses on grasping the container and aggressively tracking the predicted parabolic trajectory of the incoming object.
    
    \item \textbf{Goal Blocking:} The robot pre-positions the paddle and performs a rapid, reactive interception: 
    $\Xi_{\text{block}} = \Phi_{\mathrm{static}} \rightarrow \Phi_{\mathrm{dyn}} \rightarrow \Phi_{\mathrm{inter}}$.
    
    \item \textbf{Belt Picking:} We decompose this sequential, long-horizon task to strictly separate transport from interaction, cleanly avoiding the compounding inaccuracies typical of continuous teleoperation: 
    $\Xi_{\text{belt}} = \Phi_{\mathrm{trans}, 1} \rightarrow \Phi_{\mathrm{dyn}} \rightarrow \Phi_{\mathrm{inter}, 1} \rightarrow \Phi_{\mathrm{trans}, 2} \rightarrow \Phi_{\mathrm{inter}, 2}$. 
    Crucially, $\Phi_{\mathrm{trans}, 1}$ enforces a geometrically optimal pre-grasp posture (e.g., orienting the end-effector vertically downward) to facilitate stable visual tracking. After reactive alignment ($\Phi_{\mathrm{dyn}}$) and grasping ($\Phi_{\mathrm{inter}, 1}$), $\Phi_{\mathrm{trans}, 2}$ explicitly handles collision-free transport to the target bin, culminating in a precise release ($\Phi_{\mathrm{inter}, 2}$).
\end{itemize}

\subsubsection{Real-World Benchmark Tasks}

In real-world deployment, the composition logic is strictly refined to enforce hardware safety, boundary reachability, and cycle-bounded execution.

\begin{itemize}
    \item \textbf{Boat Loading (Reachability-Aware Tracking):} Dynamic targets often exceed the robot's strict kinematic limits. Traditional methods catastrophically fail when targets leave the workspace. To counter this, we introduce an augmented tracking phase ($\Phi_{\mathrm{dyn}}^*$) utilizing negative/boundary demonstrations: 
    $\Xi_{\text{boat}} = \Phi_{\mathrm{static}} \rightarrow \Phi_{\mathrm{dyn}}^* \rightarrow \Phi_{\mathrm{inter}} \rightarrow \Phi_{\mathrm{trans}}$. 
    Even when the boat temporarily moves out of the end-effector's reachable zone (e.g., points $A$ or $B$), $\Phi_{\mathrm{dyn}}^*$ actively projects the target to the nearest feasible boundary points (e.g., $A^*$ or $B^*$). This reachability-aware extrapolation ensures the policy aggressively pursues the target within safe kinematic bounds until a valid interaction state naturally emerges, after which it executes the placement ($\Phi_{\mathrm{inter}}$).
    
    \item \textbf{Real-World Dynamic Tapping:} To prevent physical degradation of the hardware during real-world continuous testing, the cyclic striking is strictly bounded to a finite execution ($L=3$): 
    $\Xi_{\text{tap-real}} = \Phi_{\mathrm{static}} \rightarrow (\Phi_{\mathrm{dyn}} \rightarrow \Phi_{\mathrm{inter}})^3 \rightarrow \Phi_{\mathrm{trans}}$.
    
    \item \textbf{Bottle Catching:} We utilize a highly streamlined tracking-to-grasp sequence to easily surpass human reaction limitations, ensuring minimal latency before the bottle falls: 
    $\Xi_{\text{bottle}} = \Phi_{\mathrm{dyn}} \rightarrow \Phi_{\mathrm{inter}}$.
    
    \item \textbf{Real-World Belt Picking:} To rigidly guarantee hardware safety and system repeatability between consecutive real-world trials, a mandatory terminal reset is appended: 
    $\Xi_{\text{belt-real}} = \Phi_{\mathrm{trans}, 1} \rightarrow \Phi_{\mathrm{dyn}} \rightarrow \Phi_{\mathrm{inter}, 1} \rightarrow \Phi_{\mathrm{trans}, 2} \rightarrow \Phi_{\mathrm{inter}, 2} \rightarrow \Phi_{\mathrm{trans}, 3}$. 
    Here, $\Phi_{\mathrm{trans}, 3}$ effectively neutralizes any residual kinematic momentum, safely returning the arm to its strict zero-position.
\end{itemize}

\subsection{Point Cloud Post-Processing}

Once the trajectory is edited and the augmented episode is generated, we perform post-processing on the point cloud. The primary goal is to merge point clouds from various stages, ensuring that the final scene representation is cohesive and complete.

\subsubsection{Point Cloud Merging and Cleanup}

Point clouds from different stages (e.g., Grasp, Follow, Interact, Recover) are merged to form the final scene. We ensure that redundant points are removed, and the points are uniformly distributed in the final representation.

We use a k-d tree for efficient nearest-neighbor search during the cleanup process. For each point in the scene, we check if it is too close to an object or part of the robotic arm and remove it if necessary. The cleaning is based on a threshold distance \( \delta \), which is set to 2 cm experientially.

\subsubsection{Final Point Cloud Formatting}

After cleanup, the point cloud is formatted to comply with the HDF5 data structure, where each augmented episode is stored with its associated metadata (e.g., end-effector poses, joint actions, etc.). The final point cloud is stored as a 2048-point sample for each frame, ensuring consistency across the augmented dataset.

\[
    \text{Final Point Cloud} = \{ \mathbf{p_1}, \mathbf{p_2}, \dots, \mathbf{p_{2048}} \}
\]
where each point \( \mathbf{p_i} = [x_i, y_i, z_i, r_i, g_i, b_i] \) represents the 3D coordinates and color of a point in the scene (in this case, color channels are unused).

Once the point clouds have been processed, the final augmented episodes are saved in HDF5 format. This format allows for efficient storage and retrieval of large datasets, and each augmented episode contains all the relevant data for training robotic systems.

\begin{itemize}
    \item \textbf{Episode Data:} For each episode, we store the point cloud, end-effector pose, joint action, and episode boundaries.
    \item \textbf{Metadata:} Additional metadata includes task-specific parameters and transformation matrices used during augmentation.
\end{itemize}

\section{Benchmark Details}
\label{app:benchmark_details}

\textbf{Overview}

We construct a comprehensive benchmark for dynamic manipulation tasks, encompassing both simulation and real-world environments. The simulation benchmark covers five dynamic tasks with 200 expert demonstrations each, built on the RoboTwin 2.0 platform using the SAPIEN~\cite{xiang2020sapien} physics simulator. The real-world benchmark includes four dynamic tasks with 50 demonstrations each, collected using AgileX Piper robotic arms and an Intel RealSense L515 RGB-D camera. 

\subsection{Simulation Benchmark}

Our simulation benchmark is built upon the RoboTwin 2.0 platform, using the SAPIEN physics simulator as backend. 

\subsubsection{Environment Setup}

The simulation environment consists of several key components:

\textbf{3D Assets}

We use a diverse set of 3D assets for the manipulation objects, leveraging both existing object datasets from RoboTwin and custom-designed assets. For the customized assets, we create them using Hyper3D from a single RGB image and then manually adjust the scale of the generated meshes to ensure they are of appropriate size for the tasks. We then annotate each mesh with \emph{contact points} and \emph{function points} that are relevant for the specific manipulation tasks. 

\textbf{Automatic data collection system.} The RoboTwin 2.0 platform provides a built-in data collection system that allows us to record expert demonstrations in a structured format. For each manipulation task, we design an environment script that sets up the scene with the relevant objects, initializes the robot in a predefined starting configuration, and describes the task-specific interaction logic. During trajectory planning, the system plans expert demonstration trajectories using CUROBO~\cite{sundaralingam2023curobo}, which generates smooth and feasible trajectories that achieve the task objectives. During grasping, the system iteratively tries the annotated contact points and selects the one that results in a successful grasp.

In data collection, we record multi-modal observations including RGB-D images from the head-mounted camera with 2048 points sampled from the point cloud, proprioceptive states including joint positions, end-effector poses, and gripper states, and language instructions describing the task from the task-specific templates. All data is recorded at 30\,Hz control and observation frequency and saved in HDF5 format for efficient storage and retrieval.


\subsubsection{Task descriptions.}

\textbf{Dynamic Tapping.}
The robot needs to continuously track a moving object and execute a timely tapping motion to contact the target while it is in motion.

\textbf{Belt Picking.}
The robot needs to track an object moving on a conveyor belt, grasp it within the reachable workspace, and place it into a basket.

\textbf{Object Catching.}
The robot needs to grasp a basket and estimate the trajectory of a thrown object and move a basket to the predicted landing position to catch it.

\textbf{Mole Whacking.}
The robot needs to interact with a 3$\times$3 whack-a-mole setup, where a mole appears from one hole and moves randomly toward another while leaving a visible motion trace. The robot must anticipate the mole's future position and strike it at the appropriate time.

\textbf{Goal Blocking.}
The robot needs to block an incoming ball from entering the goal by moving a handheld stick or paddle to intercept the ball's trajectory.


\subsubsection{Evaluation Metrics.}

We develop task-specific automatic evaluation criteria for each scenario to ensure objective and reproducible performance measurement. 
For Dynamic Tapping, success is determined by detecting physical contact between the robot and the moving target. 
For Belt Picking, we evaluate whether the object is successfully transported and placed within a predefined target region. 
For Object Catching, we measure whether the object is correctly captured inside the basket based on spatial inclusion. 
For Mole Whacking, success is defined by whether the robot strikes the correct target location corresponding to the mole's appearance or predicted trajectory. 
For Goal Blocking, we verify whether the ball is successfully intercepted before entering the goal region. 

All evaluations are conducted within the RoboTwin simulation environment, allowing efficient, scalable, and fully automated benchmarking. 
In addition, we record the robot's motion trajectories during task execution for further analysis of motion smoothness and dynamic tracking performance.

\subsubsection{Confidence Intervals for Simulation Evaluation}

We report 95\% confidence intervals to quantify the uncertainty of success-rate estimates from finite simulation evaluations. For each setting, we evaluate the policy over \(n=100\) trials and compute the empirical success rate \(\hat{p}=x/n\), where \(x\) is the number of successful trials. We use the Wilson score interval:
\[
\mathrm{CI}_{95\%}
=
\frac{
\hat{p} + \frac{z^2}{2n}
\pm
z\sqrt{
\frac{\hat{p}(1-\hat{p})}{n}
+
\frac{z^2}{4n^2}
}
}{
1+\frac{z^2}{n}
},
\quad z=1.96.
\]
\begin{table}[t]
  \caption{Sim data confidence intervals}
  \label{tab:sim_CI_evaluation}
  \centering
  \setlength{\tabcolsep}{3.5pt}
  \begin{tabular}{llccccc}
    \toprule
    \multicolumn{2}{c}{\textbf{Setting}} 
    & \textbf{\makecell{Dynamic\\Tapping}} 
    & \textbf{\makecell{Belt\\Picking}} 
    & \textbf{\makecell{Object\\Catching}} 
    & \textbf{\makecell{Mole\\Whacking}}  
    & \textbf{\makecell{Goal\\Blocking}} \\
    \midrule
    \multicolumn{2}{l}{1 demo}    
    & $[19\%,~36\%]$ 
    & $[0\%,~4\%]$ 
    & $[4\%,~15\%]$ 
    & $[1\%,~8\%]$ 
    & $[7\%,~20\%]$ \\

    \multicolumn{2}{l}{10 demos}  
    & $[9\%,~23\%]$ 
    & $[2\%,~11\%]$ 
    & $[44\%,~63\%]$ 
    & $[14\%,~30\%]$ 
    & $[37\%,~56\%]$ \\

    \multicolumn{2}{l}{50 demos}  
    & $[13\%,~28\%]$ 
    & $[23\%,~41\%]$ 
    & $[28\%,~47\%]$ 
    & $[53\%,~72\%]$ 
    & $[88\%,~97\%]$ \\

    \multicolumn{2}{l}{200 demos} 
    & $[45\%,~64\%]$ 
    & $[77\%,~91\%]$ 
    & $[40\%,~60\%]$ 
    & $[74\%,~89\%]$ 
    & $[79\%,~92\%]$ \\

    \midrule
    \multirow{2}{*}{\makecell{1 + aug}} 
    & (50)  
    & $[36\%,~55\%]$ 
    & $[35\%,~54\%]$  
    & $[60\%,~78\%]$ 
    & $[53\%,~72\%]$ 
    & $[80\%,~93\%]$ \\

    & (200) 
    & $[42\%,~62\%]$ 
    & $[73\%,~88\%]$  
    & $[58\%,~76\%]$ 
    & $[71\%,~87\%]$ 
    & $[79\%,~92\%]$ \\
    \bottomrule
  \end{tabular}
\end{table}

As shown in Table~\ref{tab:sim_CI_evaluation}, increasing the number of training episodes generally improves task success, though the effect varies across tasks. The augmented-data settings show that augmentation from a single expert demonstration can provide useful additional training data. 
\subsection{Real-world Benchmark}

\begin{figure*}[t]
    \centering
    \includegraphics[width=\textwidth]{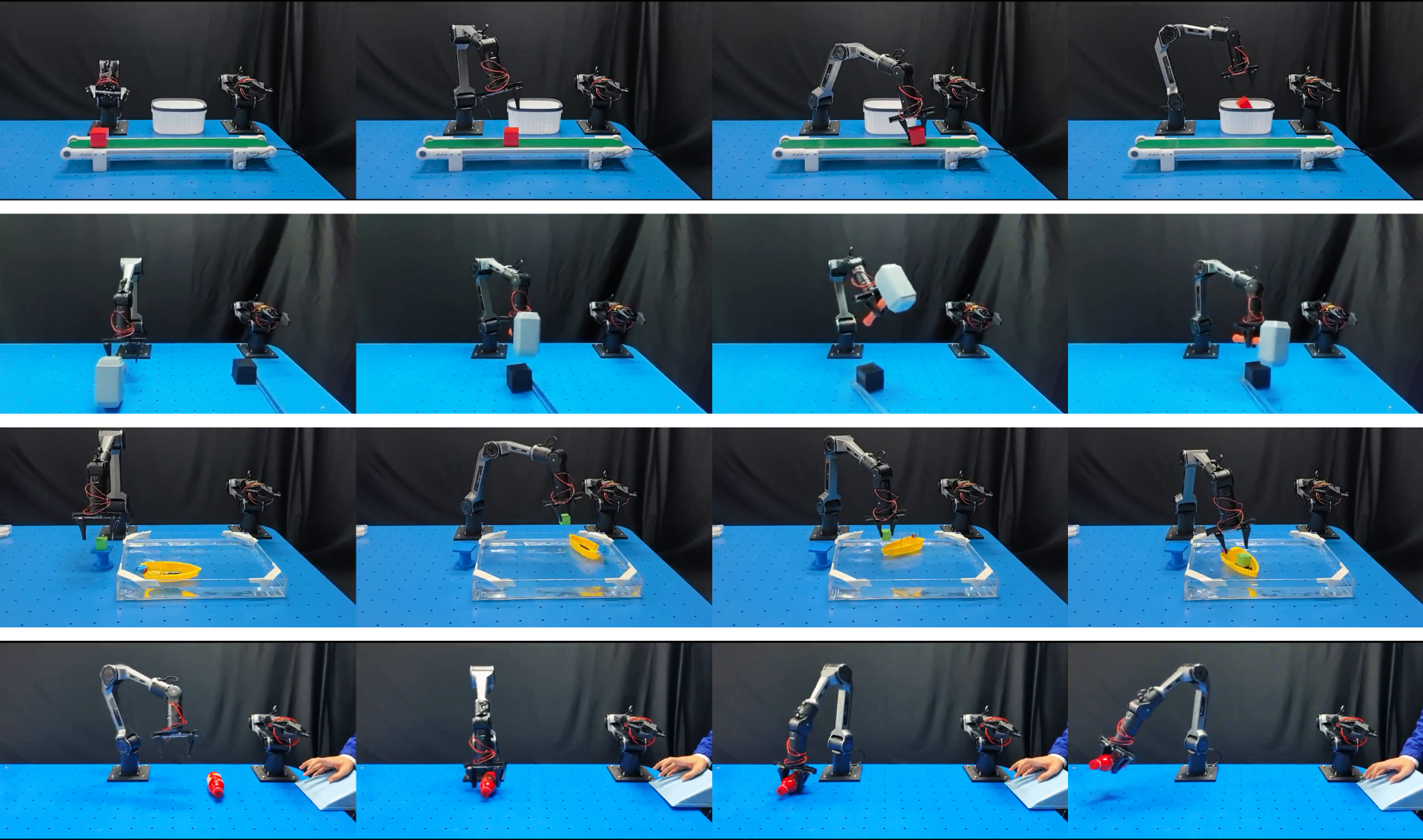}
    \caption{Visualization of autonomous policy inference and execution across four real-world dynamic manipulation tasks. From top to bottom: Belt Picking, Dynamic Tapping, Boat
    Loading, and Bottle Catching.}
    \label{fig:realworld_tasks}
\end{figure*}

Our real-world benchmark is designed to validate the effectiveness of the proposed data augmentation strategy in practical robotic scenarios.
To this end, we construct four dynamic manipulation tasks and evaluate policies trained with our augmented data against strong baselines.
Figure~\ref{fig:realworld_tasks} visualizes autonomous policy inference and execution across these tasks.
Compared to simulation, real-world dynamic tasks introduce system latency and sensing noise, making precise timing and tracking more challenging.
This places stricter requirements on the robustness and responsiveness of the learned policy, providing a more convincing evaluation of our approach.

\subsubsection{Environment Setup}

The real-world platform consists of two Piper robotic arms equipped with parallel grippers. 
Data collection is performed using a master--slave teleoperation setup with two GELLO devices. 
All demonstrations are recorded in the same format as in simulation to ensure consistency between training and deployment. 
The dataset is stored in HDF5 format and collected at 20~Hz. 

For each timestep, we record the robot joint positions and velocities, end-effector poses, and gripper states. 
In addition, we capture synchronized RGB-D observations, which are further processed into point clouds with 2048 sampled points. 
Camera intrinsic and extrinsic parameters are also stored to support geometric reasoning.

\subsubsection{Task Descriptions}

We evaluate four representative dynamic manipulation tasks in the real-world setting:

\textbf{Dynamic Tapping.}
The robot tracks a moving object and executes a precisely timed tapping motion to make contact.

\textbf{Belt Picking.}
The robot tracks an object moving on a conveyor belt, grasps it at the appropriate moment, and places it into a basket.

\textbf{Boat Loading.}
The robot grasps an object and places it onto a motorized boat moving on a water surface, requiring coordination with the boat's motion.

\textbf{Bottle Catching.}
The robot intercepts and grasps a rolling bottle before it falls off the edge of the table.

\subsubsection{Evaluation Metrics}

For each task, we perform 30 trials and report the average success rate. 
Task-specific automatic evaluation criteria are defined as follows: 
for Dynamic Tapping, success is determined by whether the robot successfully contacts the moving target; 
for Belt Picking, whether the object is grasped and placed into the basket; 
for Boat Loading, whether the object is successfully placed onto the moving boat; 
and for Bottle Catching, whether the bottle is successfully intercepted and grasped before falling. 

These metrics enable consistent and objective evaluation across all tasks.

\begin{table}[t]
  \caption{Real-world data confidence intervals}
  \label{tab:realworld_CI_evaluation}
  \centering
  \setlength{\tabcolsep}{3.5pt}
  \begin{tabular}{llcccc}
    \toprule
    \multicolumn{2}{c}{\textbf{Setting}} 
    & \textbf{\makecell{Dynamic\\Tapping}} 
    & \textbf{\makecell{Belt\\Picking}} 
    & \textbf{\makecell{Boat\\Loading}}  
    & \textbf{\makecell{Bottle\\Catching}} \\
    \midrule
    \multicolumn{2}{l}{1 demo}    
    & $[2\%,~21\%]$ 
    & $[17\%,~48\%]$ 
    & $[0\%,~11\%]$ 
    & $[0\%,~11\%]$ \\

    \multicolumn{2}{l}{50 demos}  
    & $[51\%,~83\%]$ 
    & $[52\%,~84\%]$ 
    & $[1\%,~17\%]$ 
    & $[56\%,~86\%]$ \\

    \midrule
    \multirow{2}{*}{\makecell{1 + aug}} 
    & (50)  
    & $[43\%,~76\%]$  
    & $[39\%,~73\%]$ 
    & $[12\%,~41\%]$ 
    & $[59\%,~88\%]$ \\

    & (200) 
    & $[72\%,~95\%]$  
    & $[55\%,~86\%]$ 
    & $[49\%,~81\%]$ 
    & $[66\%,~93\%]$ \\
    \bottomrule
  \end{tabular}
\end{table}

\subsubsection{Confidence Intervals for Real-world Evaluation}
As shown in Table~\ref{tab:realworld_CI_evaluation}, we report 95\% confidence intervals to quantify the uncertainty of success-rate estimates from finite real-world evaluations. For each setting, we evaluate the policy over \(n=30\) trials and compute the empirical success rate \(\hat{p}=x/n\), where \(x\) is the number of successful trials.

\section{Limitations}
\label{sec:limitations}

Despite the promising results demonstrated by DynamicManip, our framework has several limitations that point to directions for future work.

\textbf{Deformable objects.} Our data augmentation pipeline relies on mesh-enhanced geometric reconstruction that assumes rigid objects with stable geometries. For deformable or articulation-enabled objects such as cloth, ropes, or spring-loaded mechanisms, the ICP-based alignment and rigid transformation assumptions break down. Editing the trajectory of a deformable object would require modeling its evolving shape and physical parameters, which remains an open challenge.

\textbf{Rigid grasp assumption.} The current pipeline attaches the grasped object to the end-effector via a fixed relative transform \(\mathbf{T}_{\text{attach}}\) after contact. This assumption holds for tasks where the object does not slip or reorient relative to the gripper, but fails in scenarios requiring in-hand manipulation, rolling, or sliding contacts. Extending our method to handle non-rigid attachments or dynamic grasp maintenance would be a valuable extension.

\textbf{Lack of dexterous tasks.} Our benchmark tasks primarily involve parallel-jaw grippers and gross manipulation primitives such as tapping, catching, and blocking. We have not explored dexterous, multi-fingered tasks that require fine finger coordination or adaptive grasping strategies. Applying DynamicManip to dexterous manipulation would likely require augmenting the action space and rethinking the trajectory editing pipeline for multi-contact interactions.

\end{document}